\documentclass[10pt,twocolumn,letterpaper]{article}

\usepackage{iccv}
\usepackage{times}
\usepackage{epsfig}
\usepackage{graphicx}
\usepackage{amsmath}
\usepackage{amssymb}

\usepackage[accsupp]{axessibility}  
\usepackage{xcolor}
\usepackage{xspace}
\usepackage{bm}
\usepackage{multirow,multicol, makecell, booktabs}
\usepackage{wrapfig}

\usepackage{amssymb}
\usepackage{pifont}
\newcommand{\cmark}{\ding{51}}%
\newcommand{\xmark}{\ding{55}}%

\usepackage[breaklinks=true,bookmarks=false]{hyperref}

\usepackage[capitalize]{cleveref}
\crefname{section}{Sec.}{Secs.}
\Crefname{section}{Section}{Sections}
\Crefname{table}{Table}{Tables}
\crefname{table}{Tab.}{Tabs.}

\iccvfinalcopy 

\def\iccvPaperID{} 

\ificcvfinal\fi

\begin{document}


\title{VL-PET: Vision-and-Language Parameter-Efficient Tuning\\via Granularity Control}

\author{Zi-Yuan Hu$^{1,3}$ \and Yanyang Li$^1$ \and Michael R. Lyu$^1$ \and Liwei Wang$^{1,2}$\thanks{Corresponding author.} \and 
$^1$The Chinese University of Hong Kong \quad
$^2$Centre for Perceptual and Interactive Intelligence\\
$^3$Shanghai Artificial Intelligence Laboratory\\
 {\tt\small \{zyhu22,yyli21,lyu,lwwang\}@cse.cuhk.edu.hk}
}

\maketitle
\ificcvfinal\thispagestyle{empty}\fi

\begin{abstract}
As the model size of pre-trained language models (PLMs) grows rapidly, full fine-tuning becomes prohibitively expensive for model training and storage. 
In vision-and-language (VL), parameter-efficient tuning (PET) techniques are proposed to integrate modular modifications (e.g., Adapter and LoRA) into encoder-decoder PLMs. 
By tuning a small set of trainable parameters, these techniques perform on par with full fine-tuning.
However, excessive modular modifications and neglecting the functionality gap between the encoders and decoders can lead to performance degradation, while existing PET techniques (e.g., VL-Adapter) overlook these critical issues.
In this paper, we propose a 
\textbf{V}ision-and-\textbf{L}anguage \textbf{P}arameter-\textbf{E}fficient \textbf{T}uning (VL-PET) framework to impose effective control over modular modifications via a novel granularity-controlled mechanism.
Considering different granularity-controlled matrices generated by this mechanism, a variety of model-agnostic VL-PET modules can be instantiated from our framework for better efficiency and effectiveness trade-offs.
We further propose lightweight PET module designs to enhance VL alignment and modeling for the encoders and maintain text generation for the decoders.
Extensive experiments conducted on four image-text tasks and four video-text tasks demonstrate the efficiency, effectiveness and transferability of our VL-PET framework. 
In particular, our VL-PET$_\mathrm{large}$ with lightweight PET module designs significantly outperforms VL-Adapter by 2.92\% (3.41\%) and LoRA by 3.37\% (7.03\%) with BART-base (T5-base) on image-text tasks.
Furthermore, we validate the enhanced effect of employing our VL-PET designs on existing PET techniques,
enabling them to achieve significant performance improvements.
Our code is available at \href{https://github.com/HenryHZY/VL-PET}{https://github.com/HenryHZY/VL-PET}.
\end{abstract}

\begin{figure}[t]
\begin{center}
  \includegraphics[width=0.9\columnwidth]{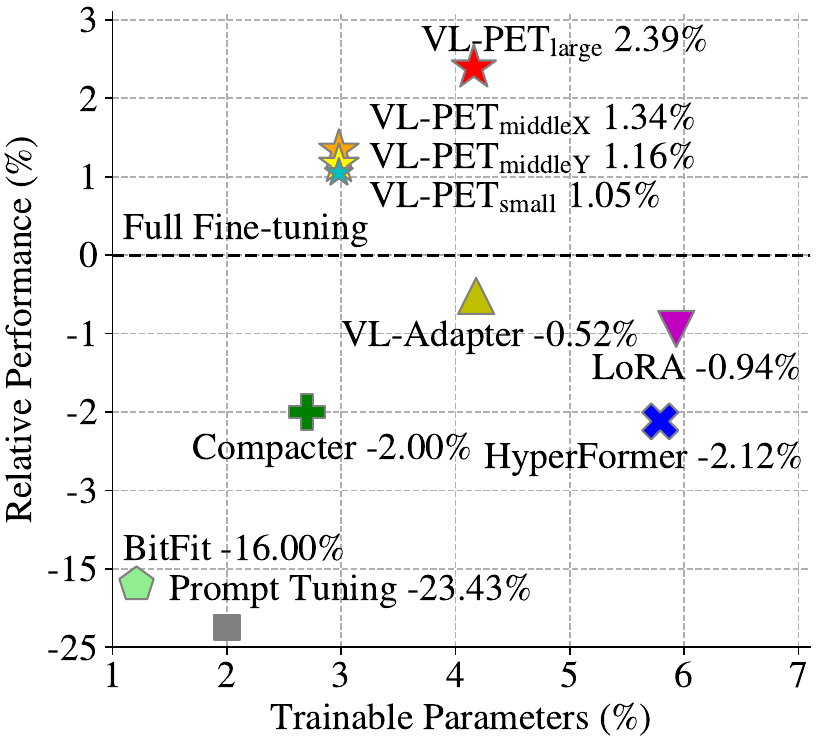}
\end{center}
   \caption{Relative average performance gain of difference PET techniques w.r.t to full fine-tuning. 
   Experiments are conducted with three seeds on four image-text tasks based on BART-base.}
\label{fig:BART_1_relative}
  \vspace{-10pt}
\end{figure}

\section{Introduction} \label{sec:intro}
Recently, the paradigm of pre-training transformer-based models on large-scale corpus and then fine-tuning them for downstream tasks has achieved great success in various domains, such as natural language processing (NLP)~\cite{vaswani2017transformer, Devlin2019bert, liu2019roberta, Lewis2020BARTDS, raffel2020t5, 2020gpt3}, computer vision (CV)~\cite{2021vit, liu2021swin, liu2021videoswin, liu2021swinv2, he2021masked, bao2022beit}, and vision-and-language (VL)~\cite{2020uniter, li2022blip, lei2021less, Li2020Hero, 2021vlt5, 2021clip, wang2022beit3}. 
However, as the model size of pre-trained language models (PLMs) and the number of tasks grow rapidly, fine-tuning the entire parameter set of PLMs (i.e., full fine-tuning) and preserving a task-specific copy of PLMs becomes prohibitively expensive for model training and storage. 

To mitigate these problems, parameter-efficient tuning (PET) techniques are proposed to save model storage space. 
As stated in~\cite{he2022UnifiedPET}, most PET techniques freeze the whole PLM backbone and integrate trainable modular modifications (i.e., additional small trainable PET modules, such as Adapter~\cite{houlsby2019parameter} and LoRA~\cite{hu2021lora}) into PLM. 
By only tuning a small set of trainable parameters, these techniques achieve performance comparable to full fine-tuning. 
Despite the significant achievements of PET in NLP~\cite{mao-etal-2022-unipelt,he2022UnifiedPET,liu2020tfew,zaken2022bitfit,karimi2021compacter,mahabadi2021hyperformer,hu2021lora,2021prefixtuning} and CV~\cite{jia2022vpt,pan2022stadapter,chen2022adaptformer,jie2023fact,chen2022convadapter,chen2022vitadapter,yu2022unifiedvpet}, 
the potential of PET in VL has not been fully explored and requires further VL-specific investigation to bridge the natural modality gap between vision and language.
In VL, most PET techniques follow NLP-specific modular modifications~\cite{manas2022mapl,ju2022vlprompt,zhou2022cocoop,zhou2022coop,khattak2022maple,yang2022ofaprompt,lu2023uniadapter}, while lacking VL-specific designs. 
Moreover, these techniques mainly focus on discriminative tasks (e.g., image-text retrieval), limiting the generalization ability of PLMs.
Although the state-of-the-art PET approach VL-Adapter~\cite{Sung2021VLAdapter} has studies challenging VL tasks, including discriminative and generative tasks (e.g., image captioning), it directly migrates those NLP-specific modular modifications without deep exploration about the most appropriate design for VL domains.

To conduct a more thorough investigation into VL-specific PET techniques, we raise and analyze \textbf{two critical issues neglected by existing PET techniques} in VL:
(1) Integrating heavy and excessive modular modifications~\cite{he2022UnifiedPET,ruder-etal-2022-modular} into PLMs can greatly affect the intermediate output of the PLMs, leading to instability and performance degradation.
Therefore, it is crucial to take measures to impose effective control over these modular modifications to achieve better performance on VL tasks.
However, state-of-the-art PET techniques (e.g., VL-Adapter) directly integrate modular modifications into PLMs without effective control.
(2) For PLMs used in VL tasks, there exists functionality gap between the encoders and  decoders~\cite{2021vlt5}. 
Specifically, the encoders focus on VL alignment and modeling, while the decoders focus on auto-regressive text generation conditioned on the visual-language representations. 
PLMs rely on the cross-attention modules inside the decoders to bridge the gap between the encoders and decoders. Therefore, it is essential to introduce tailored modular modification designs for each module, thereby enhancing their unique abilities and achieving better performance.
However, state-of-the-art PET techniques directly assign identical modular modifications to PLMs without exploring the unique ability of each PLM module, leading to suboptimal performance.

In this paper, we propose a novel \textbf{V}ision-and-\textbf{L}anguage \textbf{P}arameter-\textbf{E}fficient \textbf{T}uning (VL-PET) framework to address the above issues. 
We introduce a novel granularity-controlled mechanism to generate a granularity-controlled matrix as effective control over the modular modifications introduced by PET techniques. 
Considering different granularity control levels, a variety of granularity-controlled matrices are generated by the proposed mechanism with different trainable parameter complexities. 
With these granularity-controlled matrices and a novel multi-head modular modification, a variety of model-agnostic VL-PET modules can be instantiated from our VL-PET framework for better efficiency and effectiveness trade-offs.
Furthermore, conventional PET module designs typically integrate modular modifications (i.e., PET modules such as Adapter) into all self-attention, cross-attention, and feed-forward modules of the PLM backbones. 
Due to the unique abilities of the encoders and decoders, we propose lightweight PET module designs that facilitate suitable modular modifications integration into the encoders and decoders.
For encoders, we integrate our instantiated VL-PET modules into self-attention and feed-forward for better VL alignment and modeling.
For decoders, we only integrate our instantiated VL-PET modules into cross-attention to maintain decoder knowledge and enhance text generation. 
We further assign our instantiated VL-PET module to the value matrix inside the cross-attention, enabling refined and enhanced control over the decoders.
Subsequent experiments demonstrate that lightweight designs significantly outperform conventional designs with fewer parameters. 

Extensive experiments are conducted on four image-text tasks with BART-base~\cite{Lewis2020BARTDS}, including visual question answering (VQAv2~\cite{Goyal2017vqa} and GQA~\cite{2019gqa}), visual reasoning (NLVR$^{2}$~\cite{2019nlvr}) and image captioning (MSCOCO~\cite{2015coco}). 
As shown in~\Cref{fig:BART_1_relative}, all of the proposed VL-PET modules with lightweight PET module designs outperform the state-of-the-art PET techniques. 
In particular, VL-PET$_\mathrm{large}$ (i.e., one of our instantiated VL-PET modules) significantly outperforms VL-Adapter by 2.92\% and LoRA by 3.37\%. 
Furthermore, we transfer our model-agnostic VL-PET modules to another larger backbone (i.e., T5-base~\cite{raffel2020t5}), where the observed trends of performance improvement remain consistent with those observed in BART-base.
Our VL-PET$_\mathrm{large}$ still significantly surpasses VL-Adapter by 3.41\% and LoRA by 7.03\% in the same image-text tasks. 
However, state-of-the-art PET techniques do not show similar improvements with this larger PLM, and some techniques even exhibit performance degradation due to their heavy and excessive modular modifications integration. 
For completeness, we also transfer VL-PET modules to four video-text tasks, including video question answering (TVQA~\cite{Lei2018TVQALC} and How2QA~\cite{Li2020Hero}) and video captioning (TVC~\cite{Lei2020TVRAL} and YC2C~\cite{Zhou2018TowardsAL}). 
Comprehensive experiments and thorough ablation studies demonstrate the efficiency, effectiveness and transferability of our VL-PET framework.
Moreover, we validate the enhanced effect of employing VL-PET designs (e.g., granularity-controlled mechanism and lightweight PET module designs) on existing PET techniques (e.g., Compacter~\cite{karimi2021compacter} and VL-Adapter), enabling them to achieve significant performance improvements. 

\section{Related Work} \label{sec:related works}
\noindent\textbf{Generative Pre-trained Language Model.}
Fine-tuning pre-trained language models (PLMs) for downstream tasks has achieved great success in various domains. 
However, the architecture of PLMs also limits its applicability to downstream tasks.
PLMs with encoder-only architecture~\cite{Devlin2019bert,liu2019roberta} are effective for discriminative tasks, while PLMs with decoder-only architecture~\cite{radford2018gpt1,radford2019gpt2,2020gpt3} are better suited for generative tasks. 
PLMs with encoder-decoder architecture~\cite{vaswani2017transformer,Lewis2020BARTDS,raffel2020t5} are more generalized, as they can handle both discriminative and generative tasks.
To demonstrate the effectiveness of our VL-PET framework on challenging downstream tasks (e.g., discriminative and generative tasks), we adopt encoder-decoder generative PLMs (e.g., BART-base~\cite{Lewis2020BARTDS} and T5-base~\cite{raffel2020t5}) as our backbones.

\paragraph{Parameter-efficient Tuning.}
Parameter-efficient Tuning (PET) techniques are proposed to alleviate the exorbitant cost of model storage.
By fine-tuning PLMs with only a small set of trainable parameters, PET techniques perform on par with full fine-tuning. 
Existing PET techniques can be divided into two research categories: 
(1) As stated in~\cite{he2022UnifiedPET}, most PET techniques add new trainable parameters into PLMs (e.g., adapter-based PET techniques~\cite{houlsby2019parameter} and prompt-based PET techniques~\cite{2021prompt}); (2) Other PET techniques fine-tune a partial parameter set of the original PLMs (e.g., BitFit~\cite{zaken2022bitfit}). 
Despite the rapid development of PET techniques in NLP~\cite{mao-etal-2022-unipelt,he2022UnifiedPET,liu2020tfew,zaken2022bitfit,karimi2021compacter,mahabadi2021hyperformer,hu2021lora,2021prefixtuning} and CV~\cite{jia2022vpt,pan2022stadapter,chen2022adaptformer,jie2023fact,chen2022convadapter,chen2022vitadapter,yu2022unifiedvpet}, most PET techniques in VL are prompt-based or mainly focus on discriminative tasks~\cite{manas2022mapl,ju2022vlprompt,zhou2022cocoop,zhou2022coop,khattak2022maple,yang2022ofaprompt,lu2023uniadapter}, which limits the generalization ability of PLMs.
State-of-the-art PET techniques~\cite{Sung2021VLAdapter,Sung2022LST} focus on some challenging VL tasks with encoder-decoder generative PLMs.
Regardless of the risk of performance degradation caused by excessive modular modifications and the neglect of the unique abilities of the encoders and decoders,
VL-Adapter~\cite{Sung2021VLAdapter} directly migrates NLP-specific modular modifications without making VL-specific designs. 
In this work, we propose a VL-PET framework with a granularity-controlled mechanism, multi-head modular modifications and lightweight PET module designs to tackle these issues neglected by existing PET techniques.

\section{VL-PET Framework} \label{sec:method}
\begin{figure}[t]
\begin{center}
\includegraphics[width=\linewidth]{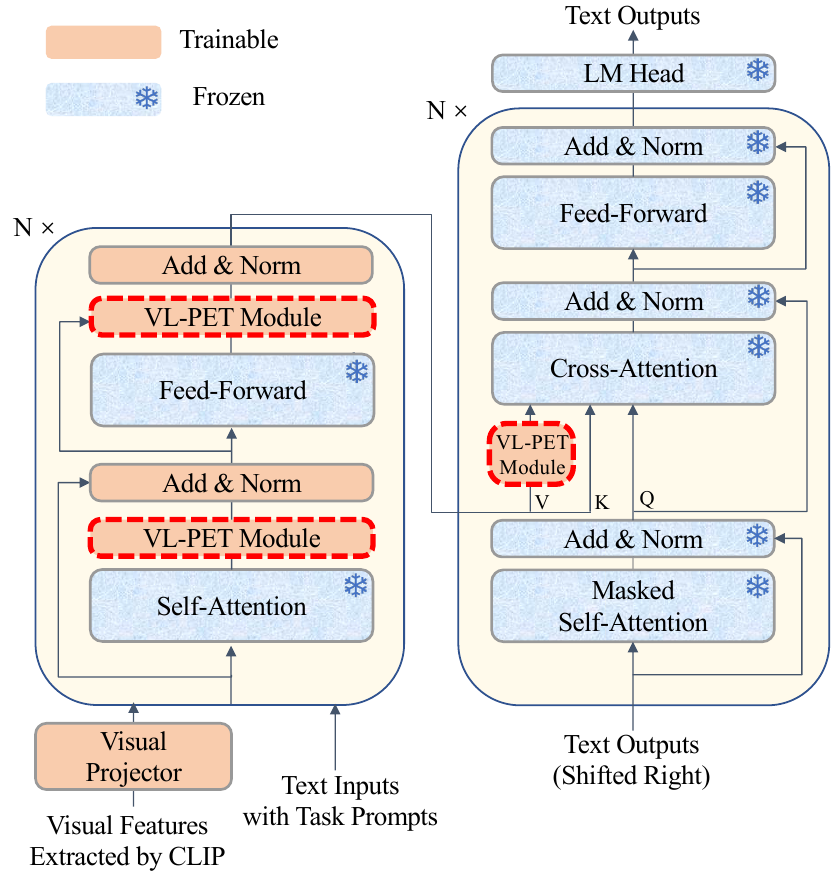}
\end{center}
   \caption{Illustration of an encoder-decoder generative pre-trained language model backbone with model-agnostic VL-PET modules and lightweight PET module designs.} 
\label{fig:model_architect}
  \vspace{-10pt}
\end{figure}

\begin{figure*}[t]
\begin{center}
\includegraphics[width=\textwidth]{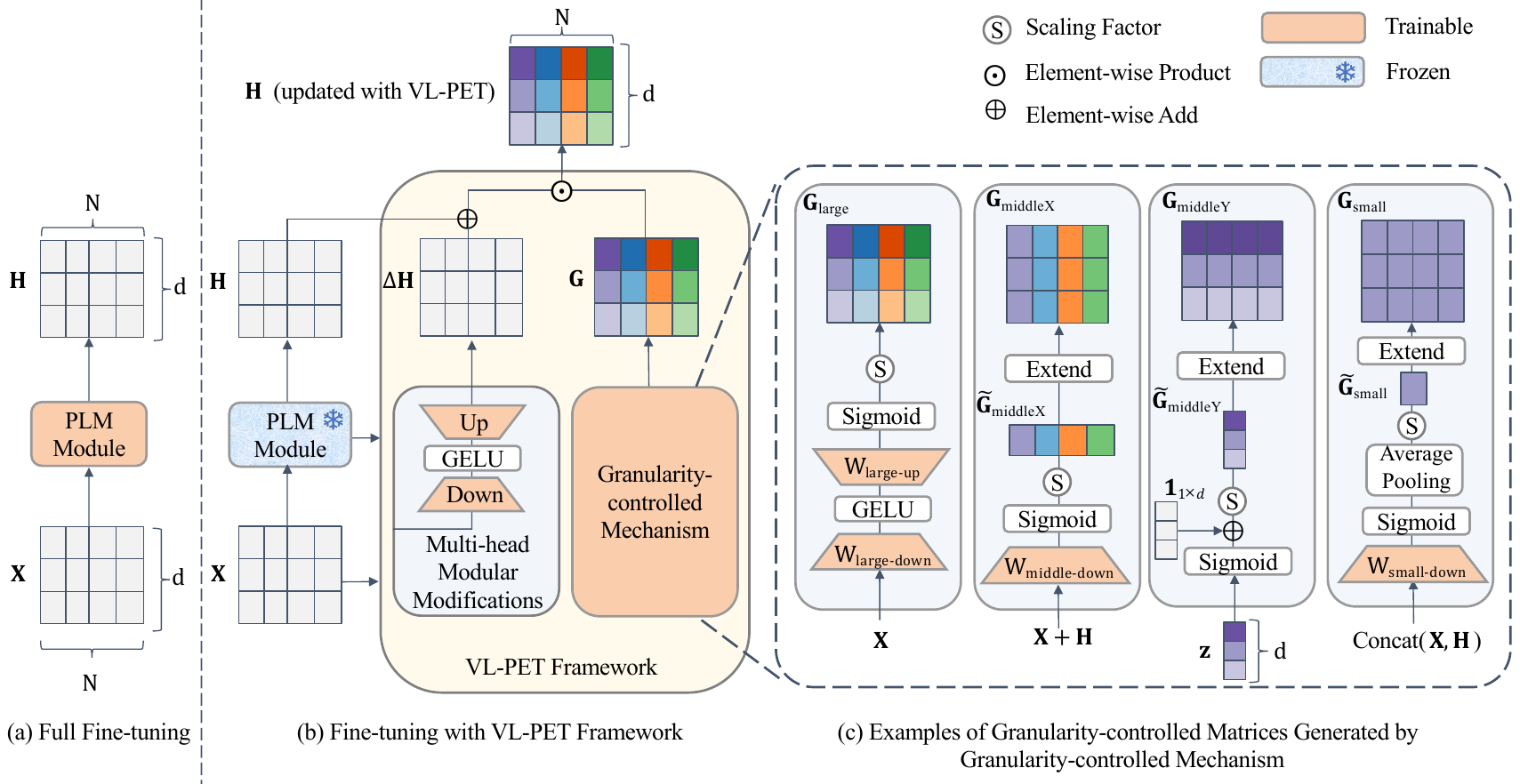}
\end{center}
   \caption{
   Comparison between full fine-tuning and fine-tuning with VL-PET framework.
   PLM module refers to a sub-module of PLMs (e.g., self-attention, feed-forward, cross-attention and value matrix of cross-attention).
   We denote $\mathbf{X}$ of length $N$ and dimension $d$ as the input of a PLM module, $\mathbf{H}$ as the output of a PLM module, $\Delta\mathbf{H}$ as a multi-head modular modification, $\bm{z}$ as a trainable vector of dimension $d$, $\mathbf{1}_{1\times d}$ as an all-one matrix
   and $\mathbf{G}$ as a granularity-controlled matrix generated by the granularity-controlled mechanism.
   }
\label{fig:model_architect2}
\end{figure*}

In this section, we propose a novel \textbf{V}ision-and-\textbf{L}anguage \textbf{P}arameter-\textbf{E}fficient \textbf{T}uning (VL-PET) framework for encoder-decoder generative PLMs. 
An illustration of our model is shown in~\cref{fig:model_architect}. 
We propose a novel granularity-controlled mechanism to generate a granularity-controlled matrix at different granularity control levels, which regulates the output of the modular modifications introduced by PET techniques. 
As shown in~\cref{fig:model_architect2}, considering different granularity control levels and a multi-head modular modification, a variety of model-agnostic VL-PET modules can be instantiated from the proposed VL-PET framework.
We further propose lightweight PET module designs to facilitate suitable VL-PET module integration into the encoders and decoders.

\subsection{Preliminary}
As stated in~\cite{he2022UnifiedPET}, most PET techniques can be attributed to introducing trainable modular modifications (e.g., Adapter~\cite{houlsby2019parameter} and LoRA~\cite{hu2021lora}) into PLMs and updating the outputs of frozen PLM modules (e.g., self-attention, feed-forward, cross-attention and value matrix of cross-attention). 
With this unified perspective, tuning with a trainable PET module can be formulated as follows:
\begin{equation}
    \mathbf{H} \leftarrow  \mathbf{H} + \Delta{\mathbf{H}}
    \label{eq:unified}
\end{equation}
where $\mathbf{H} \in \mathbb{R}^{N \times d}$ refers to an intermediate hidden state of length $N$ and dimension $d$ from a PLM module, and $\Delta\mathbf{H} \in \mathbb{R}^{N \times d}$ refers to a modular modification introduced by a PET module. 
In VL tasks, state-of-the-art methods usually migrate PET techniques~\cite{karimi2021compacter,mahabadi2021hyperformer,houlsby2019parameter} from NLP and CV without making VL-specific designs. 
Moreover, these techniques fail to impose effective control over these modular modifications, while excessive modular modifications may lead to performance degradation.
Therefore, we aim to tackle this issue with a granularity-controlled mechanism.

\subsection{Granularity-controlled Mechanism} \label{sec:granularity}
To impose effective control over the modular modifications $\Delta{\mathbf{H}}$, we propose a granularity-controlled mechanism that assigns a granularity-controlled matrix $\mathbf{G} \in \mathbb{R}^{N \times d}$ to the updating of the intermediate hidden state $\mathbf{H}$. Adopting a unified perspective from~\cref{eq:unified}, the granularity-controlled mechanism can be expressed as a unified formula:
\begin{equation}
    \mathbf{H} \leftarrow \mathbf{G} \odot (\mathbf{H} + \Delta{\mathbf{H}})
    \label{equ:granularity}
\end{equation}
where $\odot$ denotes element-wise product. 
Given the input $\mathbf{X} \in \mathbb{R}^{N \times d}$ and output $\mathbf{H}$ of a PLM module, 
a granularity-controlled matrix $\mathbf{G}$ can be generated at different granularity levels according to necessities. 
In this paper, we generate $\mathbf{G}$ into four granularity control levels (i.e., large, middleX, middleY and small).
Specifically, we directly generate a trainable matrix $\mathbf{G}_\mathrm{large} \in \mathbb{R}^{N \times d}$ at the large level.
At the middleX level, we first generate a trainable matrix $\widetilde{\mathbf{G}}_\mathrm{middleX} \in \mathbb{R}^{N \times 1}$ and then extend it to $\mathbf{G}_\mathrm{middleX} \in \mathbb{R}^{N \times d}$ without extra trainable parameters. 
Similarly, we first generate a trainable matrix $\widetilde{\mathbf{G}}_\mathrm{middleY} \in \mathbb{R}^{1 \times d}$ and $\widetilde{\mathbf{G}}_\mathrm{small} \in \mathbb{R}^{1\times 1}$ for the middleY and small, respectively, and then extend them to $\mathbf{G}_\mathrm{middleY} \in \mathbb{R}^{N \times d}$ and $\mathbf{G}_\mathrm{small} \in \mathbb{R}^{N \times d}$ without additional trainable parameters. 

Given a specific granularity-controlled matrix, a specific VL-PET module can be instantiated from our VL-PET framework.
Next, we provide one granularity-controlled matrix generation method for each granularity level, respectively, as shown in~\cref{fig:model_architect2,tab:granularity_matrix}.

\vspace{5pt}
\noindent \textbf{(1) Large Level.} 
At this level, we generate a granularity-controlled matrix $\mathbf{G}_\mathrm{large} \in \mathbb{R}^{N \times d}$ with a bottleneck architecture as follows:
\begin{equation}
    \mathbf{G}_\mathrm{large} = 
    s \cdot \sigma(\phi(\mathbf{X}\mathbf{W}_\text{large-down})\mathbf{W}_\text{large-up}) 
\end{equation}
where $\mathbf{W}_\text{large-down} \in \mathbb{R}^{d \times r}$ is a down projection layer which projects features from dimension $d$ to projected hidden dimension $r$, 
$\phi$ is a non-linear GELU function~\cite{hendrycks2016gaussian}, 
$\mathbf{W}_\text{large-up} \in \mathbb{R}^{r \times d}$ is an up projection layer,
$\sigma$ is a sigmoid function and 
$s$ is a scaling factor, which is a hyper-parameter  specialized for different PLMs. 
Therefore, the trainable parameter complexity of $\mathbf{G}_\mathrm{large}$ is $\mathcal{O}(dr)$.

\begin{table}[t]
\begin{center}
\resizebox{\columnwidth}{!}{
    \begin{tabular}{lllc}
    \toprule
    \makecell{Level} & \makecell{Trainable\\Matrix} &  \makecell{Granularity-controlled\\Matrix} & \makecell{Trainable Parameter\\Complexity} \\  
    \midrule
    Large & $\mathbf{G}_\mathrm{large} \in \mathbb{R}^{N \times d}$ & $\mathbf{G}_\mathrm{large} \in \mathbb{R}^{N \times d}$ & $\mathcal{O}(dr)$ \\
    MiddleX   & $\widetilde{\mathbf{G}}_\mathrm{middleX} \in \mathbb{R}^{N \times 1}$ & $\mathbf{G}_\mathrm{middleX} \in \mathbb{R}^{N \times d}$ & $\mathcal{O}(d)$  \\
    MiddleY   & $\widetilde{\mathbf{G}}_\mathrm{middleY} \in \mathbb{R}^{1 \times d}$ & $\mathbf{G}_\mathrm{middleY} \in \mathbb{R}^{N \times d}$ & $\mathcal{O}(d)$  \\
    Small   & $\widetilde{\mathbf{G}}_\mathrm{small} \in \mathbb{R}^{1\times 1}$ & $\mathbf{G}_\mathrm{small} \in \mathbb{R}^{N \times d}$ & $\mathcal{O}(d)$ \\
    \bottomrule
    \end{tabular}
    }
    \end{center}
    \caption{The proposed granularity-controlled matrices at different granularity control levels, which are extended from trainable matrices. 
    Trainable parameter complexity is determined by the method for generating a trainable matrix. 
    }
    \label{tab:granularity_matrix}
    \vspace{-10pt}
\end{table}

\begin{table*}[t]
\begin{center}
    \resizebox{0.95\textwidth}{!}{
    \begin{tabular}{l c c c c c c}
    \toprule
    \makecell{Method} & \makecell{Trainable\\Params (\%)} &  \makecell{VQA\\Acc. (\%)} & \makecell{GQA\\Acc. (\%)} & \makecell{NLVR$^{2}$\\Acc. (\%)} & \makecell{COCO\\ Cap. (CIDEr)} &  \makecell{Avg.}\\  
    \midrule
    \midrule
    $\textbf{Backbone: BART-base}$ \\
    \midrule
Full Fine-tuning$^\clubsuit$ & 100  & 66.88$_{0.17}$ & 56.79$_{0.41}$ & 73.66$_{0.21}$ & 112.01$_{0.93}$ & 77.33$_{0.39}$ \\
\midrule
BitFit$^\clubsuit$~\cite{zaken2022bitfit}            & 1.21 & 52.94$_{0.52}$ & 43.15$_{0.94}$ & 52.29$_{0.76}$ & 111.44$_{0.41}$ & 64.96$_{0.17}$ \\
Prompt Tuning$^\clubsuit$~\cite{2021prompt}     & 2.00    & 44.12$_{0.45}$ & 36.37$_{0.35}$ & 51.34$_{0.83}$ & 105.02$_{0.24}$ & 59.21$_{0.21}$ \\
Compacter$^\clubsuit$~\cite{karimi2021compacter}         & 2.70  & 64.63$_{0.09}$ & 52.70$_{0.24}$ & 71.11$_{0.35}$ & 114.69$_{0.42}$ & 75.78$_{0.21}$ \\
HyperFormer$^\clubsuit$~\cite{mahabadi2021hyperformer}      & 5.79 & 64.62$_{0.67}$ & 52.55$_{0.64}$ & 70.74$_{1.45}$ & 114.84$_{0.38}$ & 75.69$_{0.74}$ \\
LoRA$^\clubsuit$~\cite{hu2021lora}              & 5.93 & 65.15$_{0.16}$ & 53.66$_{0.84}$ & 72.58$_{0.73}$ & 115.01$_{0.26}$ & 76.60$_{0.32}$ \\
VL-Adapter$^\clubsuit$~\cite{Sung2021VLAdapter}        & 4.18 & \underline{65.76}$_{0.28}$ & 54.16$_{0.44}$ & \underline{73.19}$_{0.71}$ & 114.61$_{0.26}$ & 76.93$_{0.25}$ \\
    \midrule
    VL-PET$_\mathrm{small}$   & 2.98 & 65.43$_{0.06}$ & 54.03$_{0.14}$ & 72.43$_{0.22}$ & 120.68$_{0.35}$ & 78.14$_{0.11}$ \\
    VL-PET$_\mathrm{middleX}$   & 2.98 & 65.54$_{0.09}$ & \underline{54.53}$_{0.15}$ & 72.66$_{0.17}$ & \underline{120.72}$_{0.51}$ & \underline{78.37}$_{0.14}$ \\
    VL-PET$_\mathrm{middleY}$   & 2.98 & 65.36$_{0.15}$ & 53.83$_{0.39}$ & \textbf{73.43}$_{0.78}$ & 120.31$_{0.09}$ & 78.23$_{0.19}$ \\
    VL-PET$_\mathrm{large}$   & 4.16 & \textbf{66.17}$_{0.27}$ & \textbf{55.11}$_{0.17}$ & \textbf{73.43}$_{0.35}$ & \textbf{122.03$_{0.46}$} & \textbf{79.18$_{0.14}$} \\
    \midrule
    \midrule
    $\textbf{Backbone: T5-base}$ \\
    \midrule
Full Fine-tuning$^\spadesuit$ & 100  & 67.10$_{0.10}$ & 56.30$_{0.30}$ & 74.30$_{0.40}$ & 112.20$_{0.30}$ & 77.50$_{0.30}$ \\
\midrule
BitFit$^\spadesuit$~\cite{zaken2022bitfit}             & 0.83 & 55.10$_{0.20}$ & 45.50$_{0.20}$ & 51.70$_{1.10}$ & 101.20$_{0.20}$ & 63.40$_{0.10}$ \\
Prompt Tuning$^\spadesuit$~\cite{2021prompt}      & 1.26 & 47.40$_{0.70}$ & 40.60$_{0.40}$ & 51.00$_{0.40}$ & 96.10$_{0.90}$  & 58.80$_{0.60}$ \\
LoRA$^\spadesuit$~\cite{hu2021lora}               & 7.54 & 63.70$_{0.20}$ & 53.30$_{0.10}$ & 70.00$_{0.30}$ & 110.30$_{0.40}$ & 74.30$_{0.10}$ \\
VL-Adapter$^\spadesuit$~\cite{Sung2021VLAdapter}         & 7.98 & \textbf{67.10$_{0.10}$} & \underline{56.00}$_{0.40}$ & 72.70$_{0.30}$ & 111.80$_{0.10}$ & 76.90$_{0.20}$ \\
LST$^\spadesuit$~\cite{Sung2022LST}                & 7.46 & 66.50$_{0.10}$ & 55.90$_{0.10}$ & 71.60$_{0.30}$ & 113.50$_{0.30}$ & 76.90$_{0.10}$ \\
    \midrule
VL-PET$_\mathrm{small}$   & 4.51 & 65.88$_{0.31}$ & 54.96$_{1.01}$ & 72.64$_{0.09}$ & 120.05$_{0.41}$ & 78.38$_{0.37}$ \\
VL-PET$_\mathrm{middleX}$ & 4.50  & 66.63$_{0.14}$ & 55.87$_{0.25}$ & \textbf{74.11}$_{0.37}$ & \underline{120.41}$_{0.31}$ & \underline{79.26}$_{0.26}$ \\
VL-PET$_\mathrm{middleY}$ & 4.50  & 66.62$_{0.20}$ & 55.87$_{0.13}$ & 73.91$_{0.45}$ & 120.26$_{0.40}$ & 79.17$_{0.08}$ \\
VL-PET$_\mathrm{large}$   & 7.31 & \underline{66.95}$_{0.21}$ & \textbf{56.06}$_{0.21}$ & \underline{73.42}$_{0.46}$ & \textbf{121.66$_{0.06}$} & \textbf{79.52$_{0.21}$} \\
    \bottomrule
    \end{tabular}
    }
    \end{center}
    \caption{Performance on image-text tasks with different PLM backbones. 
    We report the average result with three seeds for a fair comparison, where the subscript is the standard deviation. 
    As analyzed in~\cref{sec:result1}, our special VL-PET designs (e.g., lightweight PET module designs) enable our proposed method to surpass other PET techniques in the downstream tasks (e.g., COCO Captioning).
    (\textbf{Bold} refers to the best result among all PET techniques and \underline{underline} refers to the second-best result among all PET techniques. $\clubsuit$: We reproduce the results with three seeds in~\cite{Sung2021VLAdapter}, which only provides results with one seed. $\spadesuit$: We present the results with three seeds from~\cite{Sung2022LST}.)}
    \label{tab:main table}
    \vspace{-5pt}
\end{table*}

\vspace{5pt}
\noindent \textbf{(2) MiddleX Level.} 
We first define an all-one matrix as $\mathbf{1}_{m\times n} \in \mathbb{R}^{m \times n}$. The intermediate output of the granularity-controlled mechanism at middleX level is $\widetilde{\mathbf{G}}_\mathrm{middleX} \in \mathbb{R}^{N \times 1}$, which can be calculated as follows:
\begin{equation}
    \widetilde{\mathbf{G}}_\mathrm{middleX} = 
    s \cdot \sigma((\mathbf{X}+\mathbf{H})\mathbf{W}_\text{middle-down})
\end{equation}
where $\mathbf{W}_\text{middle-down} \in \mathbb{R}^{d \times 1}$ is a down projection layer. 
Next, we use an all-one matrix $\mathbf{1}_{1\times d} \in \mathbb{R}^{1 \times d}$ to copy $\widetilde{\mathbf{G}}_\mathrm{middleX}$ $d$ times to construct $\mathbf{G}_\mathrm{middleX} \in \mathbb{R}^{N \times d}$:
\begin{equation}
    \mathbf{G}_\mathrm{middleX} = 
    \widetilde{\mathbf{G}}_\mathrm{middleX}\mathbf{1}_{1\times d}
\end{equation}
The trainable parameter complexity of $\mathbf{G}_\mathrm{middleX}$ is $\mathcal{O}(d)$.

\vspace{5pt}
\noindent \textbf{(3) MiddleY Level.} 
The granularity-controlled matrix $\mathbf{G}_\mathrm{middleY} \in \mathbb{R}^{N \times d}$ can be viewed as a variant of $\mathbf{G}_\mathrm{middleX}$ as shown in~\cref{fig:model_architect2}. 
Instead of utilizing hidden states from the PLM backbone, we adopt a trainable vector $\bm{z}\in \mathbb{R}^{1\times d}$ to calculate the intermediate output $\widetilde{\mathbf{G}}_\mathrm{middleY} \in \mathbb{R}^{1 \times d}$:
\begin{equation}
    \widetilde{\mathbf{G}}_\mathrm{middleY} = 
    s \cdot (\sigma(\bm{z}) + \mathbf{1}_{1\times d})
\end{equation}
The trainable parameter complexity of $\mathbf{G}_\mathrm{middleY}$ is $\mathcal{O}(d)$. Similarly, we extend $ \widetilde{\mathbf{G}}_\mathrm{middleY}$ to ${\mathbf{G}}_\mathrm{middleY}$ as follow:
\begin{equation}
    \mathbf{G}_\mathrm{middleY} = 
    \mathbf{1}_{N\times 1}\widetilde{\mathbf{G}}_\mathrm{middleX}
\end{equation}

\vspace{5pt}
\noindent \textbf{(4) Small Level.} 
Compared to other levels, the granularity-controlled mechanism produces the smallest intermediate output $\widetilde{\mathbf{G}}_\mathrm{small} \in \mathbb{R}^{1\times 1}$ at this level. 
We concatenate $\mathbf{X}$ and $\mathbf{H}$ along the $d$ dimension to compute $\widetilde{\mathbf{G}}_\mathrm{small}$:
\begin{equation}
    \widetilde{\mathbf{G}}_\mathrm{small} = 
    s \cdot\psi(\sigma(\text{Concat}(\mathbf{X}, \mathbf{H})\mathbf{W}_\text{small-down}))
\end{equation}
where $\mathbf{W}_\text{small-down} \in \mathbb{R}^{2d \times 1}$ is a down projection layer and $\psi$ denotes average pooling along the dimension $N$. 
Therefore, the trainable parameter complexity of this level is $\mathcal{O}(d)$. 
In the end, we expand $\widetilde{\mathbf{G}}_\mathrm{small}$ to ${\mathbf{G}}_\mathrm{small} \in \mathbb{R}^{N \times d}$:
\begin{equation}
    {\mathbf{G}}_\mathrm{small} = 
    \mathbf{1}_{N\times 1}\widetilde{\mathbf{G}}_\mathrm{small}\mathbf{1}_{1\times d}
\end{equation}

\subsection{VL-PET Module with Lightweight Designs}  \label{sec:lightweight}
\noindent \textbf{Multi-head Modular Modification.} 
Prior to introducing our lightweight PET module designs, we introduce a novel and more effective modular modification into the PLMs in a multi-head manner. 
Supposed that $N_h$ is the number of heads and $\mathbf{X}^\prime \in \mathbb{R}^{N \times d}$ is the input, a multi-head modular modification $\Delta{\mathbf{H}^\prime} \in \mathbb{R}^{N \times d}$ is defined as:
\begin{equation}
    \Delta{\mathbf{H}^\prime} 
    =  \phi(\text{Concat}(
    \mathbf{X}^\prime{\mathbf{W}_\text{down}^{(1)}}, \cdots, \mathbf{X}^\prime\mathbf{W}_\text{down}^{(N_h)}))
    \mathbf{W}_\text{up}
    \label{eqa:multihead}
\end{equation}
where $\mathbf{W}_\text{down}^{(i)}\in \mathbb{R}^{d \times \frac{r}{N_h}}$ is a down projection layer for head$_i$ and $\mathbf{W}_\text{up} \in \mathbb{R}^{r \times d}$ is a up projection layer. 
Considering a multi-head modular modification and different granularity-controlled matrices describe in~\cref{sec:granularity}, a variety of model-agnostic VL-PET modules can be instantiated from our VL-PET framework for better efficiency and effectiveness trade-offs. 

\vspace{5pt}
\noindent \textbf{Lightweight PET Module Designs.} 
Conventional PET module designs typically apply modular modifications to all self-attention, cross-attention, and feed-forward modules of the PLM backbones. 
In VL tasks, state-of-the-art PET techniques~\cite{Sung2021VLAdapter} follow these designs for encoder-decoder PLMs, neglecting the unique abilities of the encoders and decoders. 
To mitigate this issue, we propose lightweight PET module designs that facilitate suitable modular modifications integration into the encoders and decoders.
Specifically, our lightweight designs present a simple yet efficient idea that decoder PET modules should be lightweight and refined compared to encoder PET modules.

Since PLMs are trained on text-only data, adapting PLM encoders to learn unseen visual representation is crucial for VL tasks.
Therefore, we aim to enhance the visual-language alignment and modeling ability of the encoders with relatively heavy encoder PET modules.
Specifically, we integrate our instantiated VL-PET modules (utilizing $\mathbf{G}_\mathrm{large}$, $\mathbf{G}_\mathrm{middleX}$, $\mathbf{G}_\mathrm{middleY}$ or $\mathbf{G}_\mathrm{small}$) into both self-attention and feed-forward modules.
As a result, we set $\mathbf{X}^\prime$ as the output of self-attention or feed-forward modules for~\cref{eqa:multihead}.

For the decoder VL-PET modules, we want to avoid heavy and excessive modular modification in the PLM decoders as they are already good at text generation.
Since PLMs utilize cross-attention to bridge the functionality and modality gap between the encoders and decoders, we only integrate PET modules into cross-attention modules. 
Specifically, we employ our VL-PET modules (utilizing $\mathbf{G}=\mathbf{1}_{N\times d}$ for parameter-efficiency) to the value matrices of cross-attention only, enabling lightweight and refined control over the decoders. 
In this case, we set $\mathbf{X}^\prime$ as the input of value matrices (i.e., the final output of the encoders).

\begin{table}[t]
\begin{center}
    \resizebox{\columnwidth}{!}{
    \begin{tabular}{l c c c c c c}
    \toprule
    \makecell{Method} & \makecell{Trainable\\Params (\%)} &  \makecell{TVQA\\Acc. (\%)} & \makecell{How2QA\\Acc. (\%)} & \makecell{TVC\\Cap. (CIDEr)} & \makecell{YC2C\\ Cap. (CIDEr)} &  \makecell{Avg.}\\  
    \midrule

Full Fine-tuning & 100  & {77.69} & 74.79 & {50.56} & {151.71} & {88.69} \\
\midrule
BitFit           & 0.38 & 66.05 & 65.42 & 31.16 & 115.23 & 69.47 \\
Prompt Tuning    & 1.18 & 24.51 & 27.76 & 30.22 & 108.04 & 47.63 \\
Compacter        & 1.89 & 73.78 & 72.14 & 41.39 & 140.52 & 81.96 \\
LoRA             & 5.17 & 75.51 & 72.69 & 44.17 & 142.72 & 83.77 \\
VL-Adapter       & 3.39 & 77.06 & 74.73 & 46.72 & \underline{153.28} & \underline{87.95} \\
\midrule
VL-PET$_\mathrm{small}$      & 2.18 & \underline{77.69} & 74.89 & 47.92 & 150.24 & 87.69 \\
VL-PET$_\mathrm{middleY}$     & 2.17 & \textbf{77.76} & \underline{75.40} & \textbf{48.30} & 150.25 & 87.93 \\
VL-PET$_\mathrm{middleY}$   & 2.17 & 77.58 & 75.15 & \underline{47.93} & 151.13 & \underline{87.95} \\
VL-PET$_\mathrm{large}$              & 3.37 & 76.97 & \textbf{75.60} & 47.53 & \textbf{154.41} & \textbf{88.63} \\

    \bottomrule
    \end{tabular}
    }
    \end{center}
    \caption{
    Performance on video-text tasks with BART-base. 
    We report the result with one seed due to the submission limit of VALUE benchmark.
    (\textbf{Bold}: best result among all PET techniques. \underline{Underline}: second-best result among all PET techniques.)
    }
    \label{tab:video table}
\end{table}

\begin{table}[t]
\begin{center}
\resizebox{\columnwidth}{!}{
    \begin{tabular}{lcccccc}
    \toprule
    \makecell{Method} & \makecell{Params (\%)} &  \makecell{VQA (\%)} & \makecell{GQA (\%)} & \makecell{NLVR$^{2}$ (\%)} & \makecell{COCO (CIDEr)} &  \makecell{Avg.}\\  
    \midrule
    VL-PET w/o ${\mathbf{G}}$ & 2.97 & 65.22$_{0.14}$ & 53.35$_{0.39}$ & 72.65$_{0.44}$ & 120.19$_{0.68}$ & 77.85$_{0.34}$ \\
    VL-PET$_\mathrm{small}$   & 2.98 & 65.43$_{0.06}$ & 54.03$_{0.14}$ & 72.43$_{0.22}$ & 120.68$_{0.35}$ & 78.14$_{0.11}$ \\
    VL-PET$_\mathrm{middleX}$   & 2.98 & \underline{65.54}$_{0.09}$ & \underline{54.53}$_{0.15}$ & \underline{72.66}$_{0.17}$ & \underline{120.72}$_{0.51}$ & \underline{78.37}$_{0.14}$ \\
    VL-PET$_\mathrm{middleY}$   & 2.98 & 65.36$_{0.15}$ & 53.83$_{0.39}$ & \textbf{73.43$_{0.78}$} & 120.31$_{0.09}$ & 78.23$_{0.19}$ \\
    VL-PET$_\mathrm{large}$   & 4.16 & \textbf{66.17$_{0.27}$} & \textbf{55.11$_{0.17}$} & \textbf{73.43$_{0.35}$} & \textbf{122.03$_{0.46}$} & \textbf{79.18$_{0.14}$} \\

    \bottomrule
    \end{tabular}
    }
    \end{center}
    \caption{Effectiveness of granularity-controlled mechanism in BART-base.  (\textbf{Bold}: best result. \underline{Underline}: second-best result.)}
    \label{tab:Ablation_Granularity}
    \vspace{-5pt}
\end{table}

\section{Experiments}
\subsection{Experimental Settings}
\noindent \textbf{Datasets.}
In this work, we conduct experiments on four image-text downstream tasks and four video-text downstream tasks. 
Image-text tasks consist of visual question answering (VQAv2~\cite{Goyal2017vqa} and GQA~\cite{2019gqa}), visual reasoning (NLVR$^{2}$~\cite{2019nlvr}) and image captioning (MSCOCO~\cite{2015coco}). 
Video-text tasks consist of video question answering (TVQA~\cite{Lei2018TVQALC} and How2QA~\cite{Li2020Hero}) and video captioning (TVC~\cite{Lei2020TVRAL} and YC2C~\cite{Zhou2018TowardsAL}) from VALUE~\cite{li2021value} benchmark. 

\vspace{5pt}
\noindent \textbf{Baselines and Evaluations.}
Our baselines consist of state-of-the-art PET techniques, including BitFit~\cite{zaken2022bitfit}, Prompt Tuning~\cite{2021prompt}, Compacter~\cite{karimi2021compacter}, HyperFormer~\cite{mahabadi2021hyperformer}, LoRA~\cite{hu2021lora}, VL-Adapter~\cite{Sung2021VLAdapter} and LST~\cite{Sung2022LST}.
To compare these baselines with conventional PET module designs, we propose four VL-PET modules (denoted as VL-PET$_\mathrm{large}$, VL-PET$_\mathrm{middleX}$, VL-PET$_\mathrm{middleY}$ and VL-PET$_\mathrm{small}$) with lightweight PET module designs described in~\cref{sec:granularity}. 
We also include full fine-tuning (i.e., train the entire PLMs without PET modules) to facilitate a comprehensive comparison.
Following~\cite{Sung2021VLAdapter}, we adopt a trainable linear layer as the visual projector and prepend a task-specific prompt to the input sentence for each task, such as ``vqa: [Q]''.
We share the PET module for different tasks and perform multi-task learning via unified text generation~\cite{Sung2021VLAdapter} to acquire cross-task knowledge. 
We run each experiment with three seeds on image-text tasks and one seed on video-text tasks due to the submission limit of the VALUE benchmark.
The average performance of multiple tasks serves as the criterion for evaluating model performance.
To measure the efficiency of the models, we report the percentage of trainable parameters, excluding the frozen vision encoder, which is only used for offline visual feature extraction. 
We provide the implementation details in the Appendix.

\begin{table}[t]
\begin{center}
\resizebox{\columnwidth}{!}{
    \begin{tabular}{lcccccccc}
    \toprule
    \multicolumn{3}{c}{Decoder VL-PET$_\mathrm{large}$} & \multirowcell{2}{Params (\%)} &  \multirowcell{2}{VQA (\%)} & \multirowcell{2}{GQA (\%)} & \multirowcell{2}{NLVR$^{2}$  (\%)} & \multirowcell{2}{COCO (CIDEr)} &  \multirowcell{2}{Avg.}\\  
    \cmidrule{0-2}
    Self & Cross & FF & & & & & & \\
    \midrule
\cmark & \xmark &\xmark & 4.16 & 66.29$_{0.11}$ & 54.37$_{0.61}$ & 72.77$_{0.12}$ & 117.24$_{1.07}$ & 77.67$_{0.38}$ \\
 \xmark& \cmark & \xmark& 4.16 & 66.24$_{0.07}$ & 54.62$_{0.25}$ & 73.26$_{0.33}$ & 118.68$_{0.42}$ & \textbf{78.20}$_{0.25}$ \\
\xmark & \xmark &\cmark & 4.16 & 66.21$_{0.08}$ & 54.57$_{0.35}$ & 73.35$_{0.24}$ & 116.78$_{0.35}$ & 77.73$_{0.14}$ \\
\cmark & \cmark &\xmark & 4.74 & 66.66$_{0.22}$ & 55.12$_{0.28}$ & 73.16$_{0.51}$ & 116.90$_{0.98}$ & 77.96$_{0.15}$ \\
\cmark & \xmark &\cmark & 4.74 & 66.31$_{0.21}$ & 54.95$_{0.06}$ & 73.11$_{0.29}$ & 116.14$_{0.13}$ & 77.62$_{0.05}$ \\
\xmark & \cmark &\cmark & 4.74 & 66.45$_{0.24}$ & 55.13$_{0.16}$ & 73.66$_{0.23}$ & 116.94$_{0.49}$ & 78.05$_{0.04}$ \\
\cmark & \cmark &\cmark  & 5.31 & 66.57$_{0.30}$ & 54.77$_{0.26}$ & 73.54$_{0.36}$ & 115.65$_{0.46}$ & 77.63$_{0.17}$ \\
    \bottomrule
    \end{tabular}
    }
    \end{center}
    \caption{Experiments on where to integrate VL-PET$_\mathrm{large}$ into the PLM decoders. (Self, Cross and FF indicate self-attention, cross-attention and feed-forward modules of the PLM decoders. 
    \cmark: insert. 
    \xmark: not insert. 
    \textbf{Bold}: best average performance.)}
    \label{tab:decoder_pet}
\end{table}

\begin{table}[t]
\begin{center}
    \resizebox{\columnwidth}{!}{
    \begin{tabular}{l c c c c c c}
    \toprule
    \makecell{Method} & \makecell{Params (\%)} &  \makecell{VQA (\%)} & \makecell{GQA (\%)} & \makecell{NLVR$^{2}$ (\%)} & \makecell{COCO (CIDEr)} &  \makecell{Avg.}\\  
    \midrule
    Decoder VL-PET$_\mathrm{large}$ (Cross) & 4.16 & 66.24$_{0.07}$ & 54.62$_{0.25}$ & 73.26$_{0.33}$ & 118.68$_{0.42}$ & {78.20}$_{0.25}$ \\
    Decoder VL-PET$_\mathrm{large}$ (CrossK) & 4.16 & 63.25$_{0.29}$ & 53.32$_{0.25}$ & 67.14$_{1.12}$ & 114.52$_{0.54}$ & 74.55$_{0.29}$ \\
    Decoder VL-PET$_\mathrm{large}$ (CrossV)  & 4.16 & {66.17$_{0.27}$} & {55.11$_{0.17}$} & {73.43$_{0.35}$} & {122.03$_{0.46}$} & \textbf{79.18$_{0.14}$} \\
    \bottomrule
    \end{tabular}
    }
    \end{center}
    \caption{
    Experiments on how to apply VL-PET$_\mathrm{large}$ to the cross-attention modules of the PLM decoders. 
    (Cross: the whole cross-attention module. 
    CrossK: the key matrix of Cross. 
    CrossV: the value matrix of Cross. 
    \textbf{Bold}: best average performance.)
    }
    \label{tab:cross-attn}
\end{table}

\begin{table}[t]
\begin{center}
\resizebox{\columnwidth}{!}{
    \begin{tabular}{lccccccc}
    \toprule
    \multicolumn{2}{c}{LN} & \multirowcell{2}{Params (\%)} &  \multirowcell{2}{VQA (\%)} & \multirowcell{2}{GQA (\%)} & \multirowcell{2}{NLVR$^{2}$  (\%)} & \multirowcell{2}{COCO (CIDEr)} &  \multirowcell{2}{Avg.}\\  
    \cmidrule{0-1}
    Encoder LN & Decoder LN & & & & & & \\
    \midrule
 \xmark& \xmark & 4.14 & 66.17$_{0.08}$ & 54.68$_{0.16}$ & 72.42$_{0.08}$ & 121.09$_{0.42}$ & 78.59$_{0.13}$ \\
\cmark & \xmark & 4.16 & 66.17$_{0.27}$ & 55.11$_{0.17}$ & 73.43$_{0.35}$ & 122.03$_{0.46}$ & \textbf{79.18}$_{0.14}$ \\
 \xmark& \cmark & 4.16 & 66.14$_{0.24}$ & 55.06$_{0.58}$ & 73.08$_{0.22}$ & 120.09$_{0.31}$ & 78.59$_{0.19}$ \\
\cmark & \cmark & 4.18 & 66.23$_{0.12}$ & 54.60$_{0.57}$ & 72.80$_{0.26}$ & 121.18$_{0.22}$ & 78.70$_{0.13}$ \\
    \bottomrule
    \end{tabular}
    }
    \end{center}
    \caption{Effectiveness of layer normalization (LN). (\cmark: LN is trainable. \xmark: LN is frozen. 
    \textbf{Bold}: best average performance.)}
    \label{tab:ln}
    \vspace{-10pt}
\end{table}

\subsection{Main Results on Image-Text Tasks}\label{sec:result1}
To valid the efficiency, effectiveness and transferability of our VL-PET framework and model-agnostic VL-PET modules, we conduct experiments on image-text tasks with different PLM backbones (i.e., BART-base and T5-base). 

\vspace{5pt}
\noindent \textbf{Image-Text Tasks with BART-base.}
~\cref{tab:main table} has shown us the performance of full fine-tuning, state-of-the-art PET techniques and four VL-PET instantiations on image-text tasks with BART-base. 
All of our four VL-PET modules with lightweight PET module designs significantly outperform other PET techniques, as shown in~\cref{fig:BART_1_relative}, which demonstrates the efficiency and effectiveness of the proposed VL-PET framework.
Specifically, VL-PET$_\mathrm{large}$ outperforms VL-Adapter and LoRA in all downstream tasks. 
VL-PET$_\mathrm{large}$ relatively surpasses VL-Adapter by 2.92\% with comparable trainable parameters ($4.16\%<4.18\%$) and LoRA by 3.37\% with fewer trainable parameters ($4.16\%<5.93\%$). 
VL-PET$_\mathrm{small}$, VL-PET$_\mathrm{middleX}$ and VL-PET$_\mathrm{middleY}$ also surpass VL-Adapter by 1.57\%, 1.87\% and 1.69\% respectively, while utilizing far fewer trainable parameters ($2.98\% <4.18\%$). 
We observe that our VL-PET method performs comparable to or even outperform full fine-tuning, while most of the gains can be attributed to improvements in the COCO captioning task.
This phenomenon justifies the necessity of our special VL-PET designs (e.g., lightweight PET module designs) for encoders and decoders, which helps to preserve the text generation ability of the pre-trained decoders.

\begin{figure}[t]
\begin{center}
  \includegraphics[width=0.8\columnwidth]{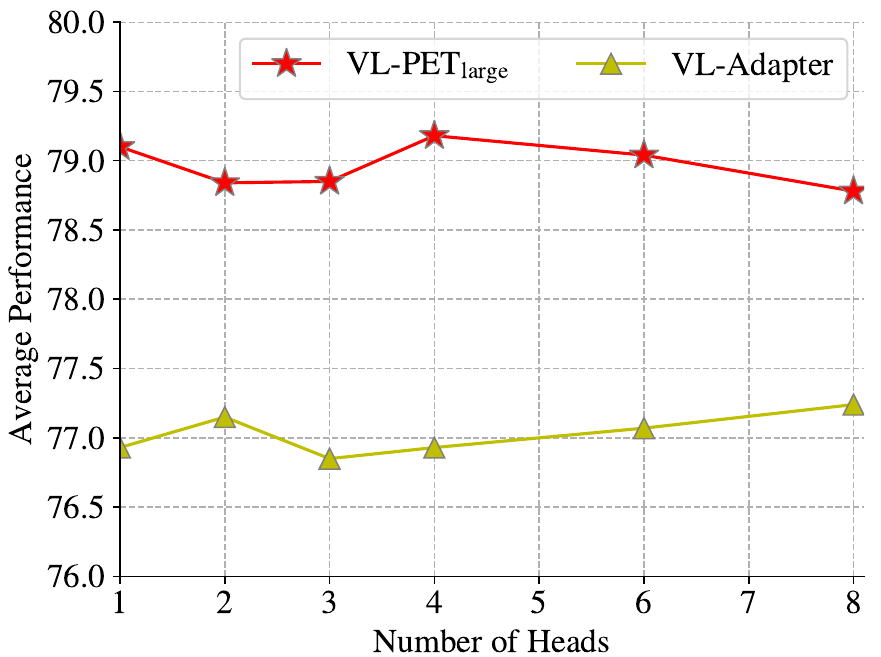}
\end{center}
   \caption{Effectiveness of the multi-head modular modification.}
\label{fig:BART_4_multihead}
\end{figure}

\begin{table}[t]
\begin{center}
    \resizebox{\columnwidth}{!}{
    \begin{tabular}{l c c c c c c}
    \toprule
    \makecell{Method} & \makecell{\#Params} &  \makecell{VQA (\%)} & \makecell{GQA (\%)} & \makecell{NLVR$^{2}$ (\%)} & \makecell{COCO (CIDEr)} &  \makecell{Avg.}\\  
    \midrule
    {VL-PET$_\mathrm{large}$ (BART-base)}\\
    Single-task Learning  & 584M & 66.13$_{0.17}$ & 53.68$_{0.18}$ & 50.36$_{1.24}$          & 121.63$_{0.72}$           & 72.95$_{0.40}$ \\
    Multi-task  Learning   & 146M & {66.17$_{0.27}$} & {55.11$_{0.17}$} & {73.43$_{0.35}$} & {122.03$_{0.46}$} & {79.18$_{0.14}$} \\
    \midrule
    {VL-PET$_\mathrm{large}$ (T5-base)}\\
    Single-task Learning  & 964M & 66.29$_{0.09}$ & 54.43$_{0.19}$          & 60.16$_{2.21}$          & 122.58$_{0.33}$           & 75.87$_{0.56}$ \\
    Multi-task  Learning  & 241M & {66.95$_{0.21}$} & {56.06$_{0.21}$} & 73.42$_{0.46}$ & {121.66$_{0.06}$} & {79.52$_{0.21}$} \\
    \bottomrule
    \end{tabular}
    }
    \end{center}
    \caption{Effectiveness of multi-task learning with different PLMs. 
    }
    \label{tab:multi-task}
    \vspace{-10pt}
\end{table}

\vspace{5pt}
\noindent \textbf{Image-Text Tasks with T5-base.}
Since the instantiated VL-PET modules are model-agnostic modules, we transfer them to another larger PLM backbone, i.e., T5-base.
As shown in~\cref{tab:main table}, the observed trends of performance improvement of VL-PET modules in T5-base remain consistent with those observed in BART-base.
All of four VL-PET modules with lightweight PET module designs significantly outperform other PET techniques. 
In particular, VL-PET$_\mathrm{large}$ outperforms LST, LoRA and VL-Adapter on most downstream tasks with fewer trainable parameters ($7.31\% < 7.46\% < 7.54\% < 7.98\%$), except for slightly lower performance on VQA compared to VL-Adapter.
Specifically, VL-PET$_\mathrm{small}$, VL-PET$_\mathrm{middleX}$, VL-PET$_\mathrm{middleY}$ and VL-PET$_\mathrm{large}$ surpasses VL-Adapter and LST by $1.92\%$, $3.07\%$, $2.95\%$ and $3.41\%$ respectively. 
They also surpasses LoRA by $5.49\%$, $6.68\%$, $6.55\%$ and $7.03\%$ respectively. 
The performances of the VL-PET modules have shown a significant improvement in a larger PLM. 
However, other PET techniques do not exhibit a similar improvement and some of them even perform was even worse than their BART-base counterparts. 
These results demonstrate the effectiveness, efficiency, and transferability of our proposed VL-PET framework.

\begin{table*}[t]
\begin{center}
\resizebox{0.95\textwidth}{!}{
    \begin{tabular}{lcccccc}
    \toprule
    \makecell{Method} & \makecell{Params (\%)} &  \makecell{VQA (\%)} & \makecell{GQA (\%)} & \makecell{NLVR$^{2}$ (\%)} & \makecell{COCO (CIDEr)} &  \makecell{Avg.}\\  
    \midrule
Compacter          & 2.70 & {64.63}$_{0.09}$ & 52.70$_{0.24}$ & {71.11}$_{0.35}$ & 114.69$_{0.42}$ & 75.78$_{0.21}$ \\
+ ${\mathbf{G}}_\mathrm{small}$ + LW & 2.08 & 64.14$_{0.05}$ & 52.84$_{0.45}$ & 71.04$_{0.54}$ & 118.77$_{0.31}$ & 76.70$_{0.08}$ \\
 + ${\mathbf{G}}_\mathrm{middleX}$ + LW           & 2.07 & 64.35$_{0.10}$ & {53.10}$_{0.73}$ & 70.57$_{0.44}$ & {119.02}$_{0.37}$ & {76.76}$_{0.17}$ \\
 + ${\mathbf{G}}_\mathrm{middleY}$ + LW           & 2.07 & 64.00$_{0.12}$ & 52.49$_{0.60}$ & 70.81$_{0.68}$ & 117.48$_{0.11}$ & 76.20$_{0.33}$ \\
 + ${\mathbf{G}}_\mathrm{large}$ + LW             & 3.28 & \textbf{65.60}$_{0.15}$ & \textbf{54.05}$_{0.30}$ & \textbf{71.66}$_{0.17}$ & \textbf{119.55}$_{0.23}$ & \textbf{77.72}$_{0.15}$ \\
\midrule
VL-Adapter        & 4.18 & {65.76}$_{0.28}$ & 54.16$_{0.44}$ & {73.19}$_{0.71}$ & 114.61$_{0.26}$ & 76.93$_{0.25}$ \\
+ ${\mathbf{G}}_\mathrm{small}$ + LW& 2.98 & 65.56$_{0.17}$ & 54.34$_{0.32}$ & 71.95$_{0.23}$ & \textbf{119.10}$_{0.25}$ & 77.74$_{0.11}$ \\
 + ${\mathbf{G}}_\mathrm{middleX}$ + LW            & 2.98 & 65.73$_{0.12}$ & {54.90}$_{0.10}$ & 73.04$_{0.50}$ & 118.34$_{0.91}$ & {78.00}$_{0.31}$ \\
 + ${\mathbf{G}}_\mathrm{middleY}$ + LW            & 2.98 & 65.67$_{0.17}$ & 54.11$_{0.29}$ & 73.18$_{0.34}$ & 117.38$_{0.52}$ & 77.58$_{0.08}$ \\
  + ${\mathbf{G}}_\mathrm{large}$ + LW              & 4.16 & \textbf{66.31}$_{0.23}$ & \textbf{55.09}$_{0.37}$ & \textbf{73.46}$_{0.37}$ & {119.05}$_{0.83}$ & \textbf{78.47}$_{0.11}$ \\
    \bottomrule
    \end{tabular}
    }
    \end{center}
    \caption{Transferring VL-PET designs to existing PET techniques.
    Experiments are conducted on image-text tasks with the BART-base backbone.
    (LW: lightweight designs and trainable encoder LNs. 
    \textbf{Bold}: the best result for different PET techniques.)}
    \label{tab:GranularityForOther}
    \vspace{-5pt}
\end{table*}

\subsection{Transfer VL-PET Modules to Video-Text Tasks}
In~\cref{tab:video table}, we also test our instantiated VL-PET modules with lightweight PET module designs on video-text tasks. 
The four VL-PET modules outperform most state-of-the-art PET techniques and attain performance comparable to full fine-tuning with the BART-base backbone. 
Concretely, VL-PET$_\mathrm{large}$ surpasses VL-Adapter by 0.77\% with comparable trainable parameters ($3.37\%<3.39\%$) and LoRA by 5.80\% with fewer trainable parameters ($3.37\%<5.17\%$). 
VL-PET$_\mathrm{small}$, VL-PET$_\mathrm{middleX}$ and VL-PET$_\mathrm{middleY}$ perform on par with VL-Adapter with fewer trainable parameters ($2.18\%<3.39\%$) and also outperform LoRA by a large margin. 
These results reveal the efficiency and effectiveness of our VL-PET framework.

\subsection{Ablation Studies}
In this section, we conduct ablation studies on image-text tasks with BART-base over three seeds by default. More experiments (e.g., visual projector, scaling factor, task prompt and weight initialization) are provided in the Appendix.

\vspace{5pt}
\noindent \textbf{Granularity-controlled Mechanism.}
In~\cref{tab:Ablation_Granularity}, we study the necessity of the proposed granularity-controlled mechanism. 
The results indicate that VL-PET modules with granularity control significantly outperform the VL-PET modules without granularity control, which demonstrates the effectiveness of our granularity-controlled mechanism.

\vspace{5pt}
\noindent \textbf{Multi-head Modular Modification.} 
We ablate the number of heads for multi-head modular modification of encoder VL-PET$_\mathrm{large}$ in~\cref{fig:BART_4_multihead} and apply this design to encoder VL-Adapter for comparison.
The best number of heads for VL-PET$_\mathrm{large}$ and VL-Adapter are 4 and 8, respectively.
The superior performance of multi-head over single-head in either VL-PET$_\mathrm{large}$ or VL-Adapter indicates the effectiveness and transferability of our multi-head modular modification.

\vspace{5pt}
\noindent \textbf{Lightweight PET Module Designs.}
\cref{tab:decoder_pet} shows our exploration of simple yet effective designs.
Compared to conventional designs, the results point out that integrating decoder PET modules in the cross-attention modules only is sufficient to achieve the best performance with fewer trainable parameters.
Subsequent experiments in~\cref{tab:cross-attn} that apply VL-PET to finer-grained PLM modules (i.e., value matrices of cross-attention) further improve the result, indicating the importance of positions of modular modifications.

\vspace{5pt}
\noindent \textbf{Layer Normalization (LN).}
Unlike VL-Adapter~\cite{Sung2021VLAdapter} which sets all LN as trainable, our experimental results in~\cref{tab:ln} indicate that utilizing trainable encoder LNs and frozen decoder LNs is a more effective strategy.

\vspace{5pt}
\noindent \textbf{Multi-Task Learning.}
Multi-task learning fine-tunes a single model for all tasks simultaneously, while single-task learning fine-tunes a single model for each task separately.
As shown in ~\cref{tab:multi-task}, multi-task learning  outperforms single-task learning on most tasks with far fewer model parameters for all tasks (BART-base: 146M$<$584M, T5-base: 241M$<$964M).
In particular, the performance on NLVR$^2$ under multi-task learning surpasses the one of single-task learning by a large margin, demonstrating that our VL-PET framework can acquire cross-task knowledge under multi-task learning.
Therefore, multi-task learning is crucial for enhancing performance and reducing model storage space.

\begin{figure*}[t]
\begin{center}
  \includegraphics[width=0.9\linewidth]{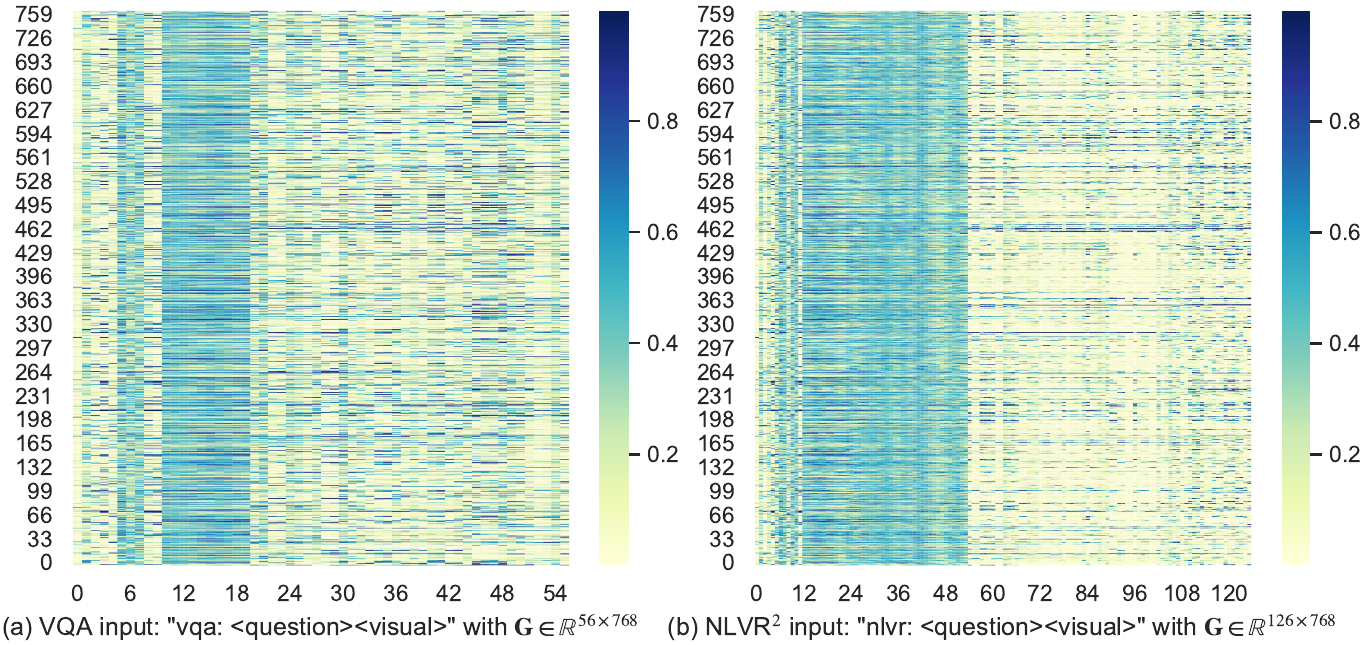}
  \vspace{-5pt}
\end{center}
   \caption{Visualizations of $\mathbf{G}_\mathrm{large}$ for two randomly picked inputs from VQAv2 and NLVR$^{2}$.}
\label{fig:visualization}
  \vspace{-10pt}
\end{figure*}

\subsection{Applying VL-PET Designs to Existing Methods}
In this section, we validate the transferability of some of our VL-PET designs to state-of-the-art PET techniques (e.g., Compacter~\cite{karimi2021compacter} and VL-Adapter~\cite{Sung2021VLAdapter}). 
As described in~\cref{sec:granularity} and~\cref{sec:lightweight}, 
we first impose effective control over the modular modifications introduced by these PET techniques.
To simplify the validation of lightweight PET module designs, we only retain the decoder PET modules in the cross-attention modules from their conventional designs.
As done in VL-PET, we similarly freeze the PLM backbone, except for the encoder LNs.
Results in~\cref{tab:GranularityForOther} demonstrate that applying our VL-PET designs to existing PET techniques leads to significant performance improvements.
In particular, Compacter and VL-Adapter with ${\mathbf{G}}_\mathrm{middleX}$ outperform their original versions by 1.29\% and 1.39\%, respectively, while utilizing fewer trainable parameters ($2.07\% < 2.70\%$ and $2.98\% < 4.18\%$). 
Compacter and VL-Adapter with ${\mathbf{G}}_\mathrm{large}$ even outperform their original performance by 2.56\% and 2.00\%, respectively.
These results again validate the universality of our VL-PET designs.

\subsection{Qualitative Analysis}
The granularity-controlled mechanism described in \cref{sec:granularity} dynamically assigns importance weights to each element in the intermediate hidden states.
To gain more insight into how it works, some visualizations are provided in~\cref{fig:visualization}, where we visualize the heatmap of $\mathbf{G}_\mathrm{large} \in \mathbb{R}^{N \times d}$ in the first encoder self-attention module of BART-base (hidden dimension $d$=768).
Given two randomly picked inputs from VQAv2 and NLVR$^{2}$, ${\mathbf{G}_\mathrm{large}}$ changes dynamically based on the inputs and thus assigns different importance weights to the hidden states.
For some text tokens, large weights are densely assigned to almost all of their elements. 
For other tokens (especially vision tokens), large weights are sparsely distributed on their elements. 
Such learned weight assignment strategies attest that our granularity-controlled mechanism is a non-trivial method.

\section{Conclusion} \label{sec:conclusion}
In this paper, we analyze and tackle some critical issues overlooked by existing PET techniques on VL tasks (e.g., effective control over modular modifications, the encoder-decoder connections and the unique abilities of encoders and decoders).
We propose a novel VL-PET framework to effectively control the modular modifications introduced by PET techniques via a granularity-controlled mechanism. 
Considering different granularity control levels, multi-head modular modifications and lightweight PET module designs, a variety of model-agnostic VL-PET modules can be instantiated from the proposed VL-PET framework for better efficiency and effectiveness trade-offs.
Extensive experiments conducted on image-text and video-text tasks demonstrate the efficiency, effectiveness and transferability of our VL-PET framework. 
Moreover, we validate the universality of our VL-PET designs (e.g., granularity-controlled mechanism and lightweight PET module designs) by transferring them to existing PET techniques, enabling them to achieve significant performance improvements. 
Although our work focuses on VL tasks, the ideas and designs presented in this work (e.g., granularity-controlled mechanism, multi-head modular modifications and lightweight PET module designs) have the potential to be applied to other domains (e.g., NLP and CV).

\section*{Acknowledgments}
This research was partially funded by the Centre for Perceptual and Interactive Intelligence (CPII) Ltd under the Innovation and Technology Commission (ITC)’s InnoHK. 
Liwei Wang is a Principal Investigator of CPII under the InnoHK. 
This work was also partially supported by the UGC under Research Matching Grant Scheme and Direct Grant at CUHK, and Research Grants Council of the Hong Kong Special Administrative Region, China (No. CUHK 14206921 of the General Research Fund).

{\small
\bibliographystyle{ieee_fullname}
\bibliography{egbib}
}

\appendix
\newpage
\section{Dataset Statistics}
For image-text tasks, we report the performance on Karpathy test/test-dev/test-P/Karpathy test split for VQA/GQA/NLVR$^2$/MSCOCO. 
For video-text tasks, we report the performance on test-public split for TVQA, How2QA, TVC and YC2C from VALUE benchmark~\cite{li2021value}.
The detailed statistics of image-text and video-text datasets are shown in~\cref{tab:data_image} and~\cref{tab:data_video}, respectively.

\section{Implementation Details}
Combining a pre-trained vision model with a generative PLM, we perform multi-task learning via unified text generation. 
Following~\cite{2021vlt5}, we use CLIP~\cite{2021clip} as our frozen vision encoder for offline visual feature extraction, and BART-base~\cite{Lewis2020BARTDS} and T5-base~\cite{raffel2020t5} as our generative PLM backbones for parameter-efficient tuning. 
A trainable visual projector (i.e., a trainable linear layer) is utilized to project the visual features into the dimension space of text embeddings. 
The projected visual features and text embeddings are subsequently concatenated as input into our PLM backbone. 
For a fair experimental comparison, we follow the setting of existing PET techniques~\cite{Sung2021VLAdapter, Sung2022LST} to adopt a shared-weight PET module for all downstream tasks and use CLIP-ResNet101~\cite{2021clip,2016resnet} as our vision encoder for image-text tasks and CLIP (ViT-B/32)~\cite{2021clip,2021vit} for video-text tasks. 
Specifically, images are resized to $224 \times 224$ at first, and then $6 \times 6$ grid features are extracted by the last convolutional layer with an adaptive maximum pooling.
Video features are extracted offline at a fixed frame rate (one frame per second) with a maximum length of 64.

We train the models for 20 epochs for both image-text tasks and video-text tasks. 
The batch size is set as 500 for BART-base on image-text tasks, 300 for T5-base on image-text tasks, and 50 for BART-base on video-text tasks. 
We use AdamW~\cite{loshchilov2017adamw} as our optimizer to train the models and apply a linear decay scheduler with a warmup ratio of 0.1. 
We evaluate the last checkpoints of models. 
All experiments are conducted on one A100 GPU (80G memory). 
The average training time is 16 hours for BART-base on image-text tasks, 18 hours for T5-base on image-text tasks, and 18 hours for BART-base on video-text tasks.

\section{Hyper-Parameters}
For image-text tasks with the BART-base backbone,~\cite{Sung2021VLAdapter} conduct a detailed hyper-parameter search to find the optimal settings for the baselines. However, it only reports experimental results of one random seed. 
To strengthen the reliability of these experiments, we summarize statistics of three seeds with the hyper-parameters listed in~\cref{tab:hyper1}.

\begin{table}[t]
\begin{center}
    \resizebox{\columnwidth}{!}{
    \begin{tabular}{llccc}
    \toprule
     \multirowcell{2}{Dataset} & 
     \multicolumn{3}{c}{Statistics (\#images / \#QA pairs, \#captions)} \\
    & \makecell{Train} & \makecell{Validation} & \makecell{Test}\\
    \midrule
    
     VQA \cite{Goyal2017vqa} & 113.2K/605.1K & 5.0K/26.7K & 5.0K/26.3K  \\
     GQA \cite{2019gqa}  & 72.1K/943.0K & 10.2K/132.1K & 398/12.6K\\
     NLVR$^2$ \cite{2019nlvr} & 103.2K/86.4K & 8.1K/7.0K & 8.1K/7.0K  \\
     MSCOCO \cite{2015coco} & 113.2K/566.8K & 5.0K/5.0K & 5.0K/5.0K \\
    \bottomrule
    \end{tabular}
    }
    \end{center}
    \caption{The statistics of four image-text datasets.}
    \label{tab:data_image}
\end{table}

\begin{table}[t]
\begin{center}
    \resizebox{\columnwidth}{!}{
    \begin{tabular}{llccc}
    \toprule
     \multirowcell{2}{Dataset} & 
     \multicolumn{3}{c}{Statistics (\#videos / \#QA pairs, \#captions)} \\
    & \makecell{Train} & \makecell{Validation} & \makecell{Test}\\
    \midrule
    TVQA \cite{Lei2018TVQALC} & 17.4K/122.0K & 2.2K/15.3K &  2.2K/15.3K \\
    How2QA \cite{Li2020Hero} & 24.5K/34.2K & 3.1K/3.1K & 3.1K/3.1K \\
    TVC \cite{Lei2020TVRAL} & 17.4K/86.7K & 10.8K/43.6K & 10.8K/43.6K \\
    YC2C \cite{Zhou2018TowardsAL} &  10.3K/10.3K & 3.5K/3.5K & 1.6K/1.6K \\
    \bottomrule
    \end{tabular}
    }
    \end{center}
    \caption{The statistics of four video-text datasets.}
    \label{tab:data_video}
\end{table}

\begin{table}[t]
\begin{center}
\resizebox{\columnwidth}{!}
{\begin{tabular}{llll}
\toprule
Task & Visual Input & Text Input with Task Prompt & Text Output  \\
\midrule
VQA & image features & vqa: \texttt{[Q]} & \texttt{[A]} \\
GQA & image features & gqa: \texttt{[Q]} & \texttt{[A]} \\
NLVR & image features & nlvr: \texttt{[text]} & true/false \\
MSCOCO & image features & caption: & \texttt{[caption]} \\
\midrule
TVQA & video features & tvqa: \texttt{[Q]} & \texttt{[A]} \\
How2QA & video features & how2qa: \texttt{[Q]} & \texttt{[A]} \\
TVC & video features & tvc: & \texttt{[caption]} \\
YC2C & video features & yc2c: & \texttt{[caption]} \\
\bottomrule
\end{tabular}
}
\end{center}
\caption{Input-output formats with task prompts from~\cite{2021vlt5,Sung2021VLAdapter}. \texttt{[Q]} denotes question, \texttt{[A]} denotes answer, \texttt{[text]} denotes text and \texttt{[caption]} denotes captioning results.}
\label{table:task_prompt}
\end{table}


For image-text tasks with the T5-base backbone, we borrow the results of three seeds from~\cite{Sung2022LST}. The hyper-parameters of the baselines are listed in~\cref{tab:hyper2}.

For video-text tasks with the BART-base backbone, we show the results of only one seed due to the submission limit of the VALUE benchmark. The hyper-parameters are listed in~\cref{tab:hyper3}.

\begin{table*}[!t]
\begin{center}
    \resizebox{\textwidth}{!}{
    \begin{tabular}{l c c c l}
    \toprule
    \makecell{Method} & \makecell{Learning Rate} & \makecell{Batch Size} & \makecell{Epoch} & \makecell{Other Hyper-Parameters} \\
    \midrule
     Full Fine-tuning & $1 \times 10^{-4}$ & 500 & 20 & - \\
     \midrule
     BitFit~\cite{zaken2022bitfit} & $1 \times 10^{-3}$ & 500 & 20 &- \\
    Prompt Tuning~\cite{2021prompt} & $1 \times 10^{-3}$ & 500 & 20 & prompt length $N_p = 40$, prompt dimension $d_m = 800$ \\    
    Compacter~\cite{karimi2021compacter} & $1 \times 10^{-3}$ & 500 & 20 &hidden dimension $d = 96$, Kronecker products $k = 2$ \\
     Hyperformer~\cite{mahabadi2021hyperformer} & $1 \times 10^{-3}$ & 500 & 20 & hidden dimension $d = 96$, task dimension $d_p = 8$ \\
     LoRA~\cite{hu2021lora}  & $1 \times 10^{-3}$ & 500 & 20 & hidden dimension $d = 128$\\
    VL-Adapter~\cite{Sung2021VLAdapter} & $1 \times 10^{-3}$ & 500 & 20 & hidden dimension $d = 96$ \\
     \midrule
    VL-PET$_\mathrm{small}$  & $1 \times 10^{-3}$ & 500 & 20 &  \multirow{4}{*}{encoders: $r = 96$, $s=1.0$, $N_h=4$; decoders: $r = 96$, $s=1.0$, $N_h=1$}  \\
    VL-PET$_\mathrm{middleX}$   & $1 \times 10^{-3}$ & 500 & 20 &  \\
    VL-PET$_\mathrm{middleY}$  & $1 \times 10^{-3}$ & 500 & 20 &  \\
    VL-PET$_\mathrm{large}$   & $1 \times 10^{-3}$ & 500 & 20 &  \\
    \bottomrule
    \end{tabular}
    }
    \end{center}
    \caption{
    Hyper-parameters for image-text tasks with BART-base backbone. We follow the baseline settings of~\cite{Sung2021VLAdapter}.
    }
    \label{tab:hyper1}
\end{table*}

\begin{table*}[!t]
\begin{center}
    \resizebox{\textwidth}{!}{
    \begin{tabular}{l c c c l}
    \toprule
    \makecell{Method} & \makecell{Learning Rate} & \makecell{Batch size} & \makecell{Epoch} & \makecell{Other hyper-parameters} \\
    \midrule
     Full Fine-tuning & $3 \times 10^{-4}$ & 300 & 20 & - \\
     \midrule
     BitFit~\cite{zaken2022bitfit} & $3 \times 10^{-3}$ & 300 & 20 & - \\
    Prompt Tuning~\cite{2021prompt} & $3 \times 10^{-3}$ & 300 & 20 &  prompt length $N_p = 40$, prompt dimension $d_m = 800$ \\    
     LoRA~\cite{hu2021lora}  & $3 \times 10^{-4}$ & 300 & 20 & hidden dimension=150\\
    VL-Adapter~\cite{Sung2021VLAdapter} & $3 \times 10^{-4}$ & 300 & 20  & hidden dimension $d = 192$ \\
    LST~\cite{Sung2022LST} & $3 \times 10^{-3}$ & 300 & 20  & r=4; all layers in side networks are kept \\
     \midrule
    VL-PET$_\mathrm{small}$  & $3 \times 10^{-4}$ & 300 & 20 & \multirow{4}{*}{encoders: $r = 192$, $s=0.3$, $N_h=4$; decoders: $r = 96$, $s=1.0$, $N_h=1$}  \\
    VL-PET$_\mathrm{middleX}$   & $3 \times 10^{-4}$ & 300 & 20 &  \\
    VL-PET$_\mathrm{middleY}$  & $3 \times 10^{-4}$ & 300 & 20 &  \\
    VL-PET$_\mathrm{large}$   & $3 \times 10^{-4}$ & 300 & 20 &  \\
    \bottomrule
    \end{tabular}
    }
    \end{center}
    \caption{
    Hyper-parameters for image-text tasks with T5-base backbone. We follow the baseline settings of~\cite{Sung2022LST}.
    }
    \label{tab:hyper2}
\end{table*}

\begin{table*}[!t]
\begin{center}
    \resizebox{\textwidth}{!}{
    \begin{tabular}{l c c c l}
    \toprule
    \makecell{Method} & \makecell{Learning Rate} & \makecell{Batch size} & \makecell{Epoch} & \makecell{Other hyper-parameters} \\
    \midrule
     Full Fine-tuning & $1 \times 10^{-5}$ & 50  & 20 & - \\
     \midrule
     BitFit~\cite{zaken2022bitfit} & $1 \times 10^{-4}$ & 50  & 20 & - \\
    Prompt Tuning~\cite{2021prompt} & $1 \times 10^{-4}$ & 50  & 20 &  prompt length $N_p = 40$, prompt dimension $d_m = 800$ \\    
    Compacter~\cite{karimi2021compacter} & $1 \times 10^{-4}$ & 50  & 20 & hidden dimension $d = 96$, Kronecker products $k = 2$ \\
     LoRA~\cite{hu2021lora}  & $1 \times 10^{-4}$ & 50  & 20 & hidden dimension $d = 128$\\
    VL-Adapter~\cite{Sung2021VLAdapter} & $1 \times 10^{-4}$ & 50  & 20 & hidden dimension $d = 96$ \\
     \midrule
    VL-PET$_\mathrm{small}$  & $7 \times 10^{-4}$ & 50  & 20 & \multirow{4}{*}{encoders: $r = 96$, $s=1.0$, $N_h=4$; decoders: $r = 96$, $s=1.0$, $N_h=1$}  \\
    VL-PET$_\mathrm{middleX}$    & $7 \times 10^{-4}$ & 50  & 20 &  \\
    VL-PET$_\mathrm{middleY}$   & $7 \times 10^{-4}$ & 50  & 20 &  \\
    VL-PET$_\mathrm{large}$    & $7 \times 10^{-4}$ & 50  & 20 &  \\
    \bottomrule
    \end{tabular}
    }
    \end{center}
    \caption{
    Hyper-parameters for video-text tasks based on BART-base backbone.
    }
    \label{tab:hyper3}
\end{table*}

\begin{table*}[t]
\begin{center}
    \resizebox{0.95\textwidth}{!}{
    \begin{tabular}{l c c c c c c}
    \toprule
    \makecell{Input} & \makecell{Params (\%)} &  \makecell{VQA (\%)} & \makecell{GQA (\%)} & \makecell{NLVR$^{2}$ (\%)} & \makecell{COCO (CIDEr)} &  \makecell{Avg.}\\  
    \midrule
    {VL-PET$_\mathrm{large}$ (BART-base)}\\
    T & 3.10 & 44.80$_{0.29}$ & 40.19$_{0.16}$ & 51.09$_{0.07}$ & 6.66$_{0.41}$   & 35.69$_{0.03}$ \\
    T + Noise  & 4.16 & 44.79$_{0.10}$ & 40.49$_{0.16}$ & 51.09$_{0.22}$ & 6.57$_{0.98}$   & 35.74$_{0.16}$ \\
    T + Frozen V & 3.07 & 62.62$_{0.50}$ & 51.34$_{0.67}$ & 67.77$_{0.39}$ & 117.58$_{1.90}$ & 74.83$_{0.72}$ \\
    T + Trainable V & 4.16 & {66.17$_{0.27}$} & {55.11$_{0.17}$} & {73.43$_{0.35}$} & {122.03$_{0.46}$} & {79.18$_{0.14}$} \\
    \midrule
    {VL-PET$_\mathrm{large}$ (T5-base)}\\
    T & 6.70 & 44.90$_{0.12}$ & 40.62$_{0.11}$ & 51.21$_{0.09}$ & 4.81$_{0.77}$   & 35.39$_{0.19}$ \\
    T + Noise & 7.31 & 44.85$_{0.18}$ & 40.36$_{0.02}$ & 51.22$_{0.09}$ & 6.45$_{1.81}$   & 35.72$_{0.42}$ \\
    T + Frozen V & 6.65 & 64.30$_{0.20}$ & 53.18$_{0.55}$ & 69.90$_{0.68}$ & 117.90$_{0.37}$ & 76.32$_{0.44}$ \\
    T + Trainable V   & 7.31 & {66.95$_{0.21}$} & {56.06$_{0.21}$} & 73.42$_{0.46}$ & {121.66$_{0.06}$} & {79.52$_{0.21}$} \\
    \bottomrule
    \end{tabular}
    }
    \end{center}
    \caption{Effectiveness of visual features and visual projector. (T: text. Noise: noise features. V: visual projector.)}
    \label{tab:visual features}
\end{table*}

\begin{table*}[t]
\begin{center}
\resizebox{0.95\textwidth}{!}{
    \begin{tabular}{lcccccc}
    \toprule
    \makecell{Method} & \makecell{Params (\%)} &  \makecell{VQA (\%)} & \makecell{GQA (\%)} & \makecell{NLVR$^{2}$ (\%)} & \makecell{COCO (CIDEr)} &  \makecell{Avg.}\\  
    \midrule
    VL-PET$_\mathrm{large}$ with $\Delta{\mathbf{H}}^{\mathrm{vis}}$  & 3.29 & 65.46$_{0.15}$ & 53.94$_{0.27}$ & 72.89$_{0.32}$ & 120.15$_{0.68}$ & 78.11$_{0.13}$ \\
    VL-PET$_\mathrm{large}$ with scaled up $\Delta{\mathbf{H}}^{\mathrm{vis}}$  & 3.47 & 65.57$_{0.03}$ & 53.70$_{0.70}$ & 73.15$_{0.41}$ & 120.00$_{0.24}$ & 78.11$_{0.12}$ \\
    VL-PET$_\mathrm{large}$ with $\Delta{\mathbf{H}}^{\mathrm{vis}}$ and $\mathbf{G}_\mathrm{large}^{\mathrm{vis}}$  & 3.47 & 65.50$_{0.13}$ & 54.03$_{0.44}$ & 73.35$_{0.18}$ & 120.71$_{0.43}$ & 78.40$_{0.03}$ \\
    VL-PET$_\mathrm{large}$   & 4.16 & {66.17$_{0.27}$} & {55.11$_{0.17}$} & {73.43$_{0.35}$} & {122.03$_{0.46}$} & {79.18$_{0.14}$} \\

    \bottomrule
    \end{tabular}
    }
    \end{center}
    \caption{Decomposing the visual projector on BART-base.}
    \label{tab:visual_projector}
\end{table*}

\begin{table*}[!t]
\begin{center}
\resizebox{0.95\textwidth}{!}{
    \begin{tabular}{llcccccc}
    \toprule
    \makecell{Method} & \makecell{Updating Formula} &\makecell{Params (\%)} &  \makecell{VQA (\%)} & \makecell{GQA (\%)} & \makecell{NLVR$^{2}$ (\%)} & \makecell{COCO (CIDEr)} &  \makecell{Avg.}\\  
    \midrule
    VL-PET w/o granularity control & $\mathbf{H} \leftarrow  \mathbf{H} + \Delta{\mathbf{H}}$ & 2.97 & 65.22$_{0.14}$ & 53.35$_{0.39}$ & 72.65$_{0.44}$ & 120.19$_{0.68}$ & 77.85$_{0.34}$ \\
    VL-PET$_\mathrm{add}$  & $\mathbf{H} \leftarrow  \mathbf{H} + \Delta{\mathbf{H}} + \mathbf{G}_\mathrm{large}$ & 4.16 & 65.10$_{0.05}$ & 53.26$_{0.26}$ & 71.85$_{0.19}$ & 121.26$_{1.19}$ & 77.87$_{0.26}$ \\
    VL-PET$_\mathrm{large}$ & $\mathbf{H} \leftarrow \mathbf{G}_\mathrm{large} \odot (\mathbf{H} + \Delta{\mathbf{H}})$ & 4.16 & {66.17$_{0.27}$} & {55.11$_{0.17}$} & {73.43$_{0.35}$} & {122.03$_{0.46}$} & {79.18$_{0.14}$} \\

    \bottomrule
    \end{tabular}
    }
    \end{center}
    \caption{Performance improvement effect of granularity-controlled mechanism.}
    \label{tab:granularity_add}
\end{table*}

\section{Effectiveness of Visual Features and Trainable Visual Projector}
In this section, we investigate the effectiveness of the visual features and a trainable visual projector on VL tasks. 
We denote T as text, Noise as noise features sampled from a uniform distribution on the interval from 0 to 1, Trainable V as visual features with a trainable visual projector and Frozen V as visual features with a frozen visual projector.
We replace the standard model input (i.e., T + Trainable V) with T (text-only), T + Noise, and T + Frozen V.
In~\cref{tab:visual features}, T + Trainable V outperforms other inputs by a large margin, demonstrating the effectiveness of visual features and the trainable visual projector, as well as the importance of VL alignment and modeling on VL tasks.

In~\cref{tab:visual_projector}, we also decompose the visual projector using multi-head modular modifications (denoted as $\Delta{\mathbf{H}}^{\mathrm{vis}}$, $r$=96, $N_h$=4) and the granularity control at large level (denoted as $\mathbf{G}_\mathrm{large}^{\mathrm{vis}}$). 
Due to the different dimensions of visual and text features, we have to remove the residual connection for $\Delta{\mathbf{H}}^{\mathrm{vis}}$.
The results indicate that simply scaling up $\Delta{\mathbf{H}}^{\mathrm{vis}}$ (from $r$=96 to $r$=192) dose not effectively improve the performance, whereas introducing $\mathbf{G}_\mathrm{large}^{\mathrm{vis}}$ into $\Delta{\mathbf{H}}^{\mathrm{vis}}$ is more effective.
Although decomposing the visual projector provides us more efficiency and effectiveness trade-offs, we do not include it in the main paper for a fair competition with the existing PET techniques (e.g., VL-Adapter).

\begin{figure}[t]
\begin{center}
  \includegraphics[width=0.9\columnwidth]{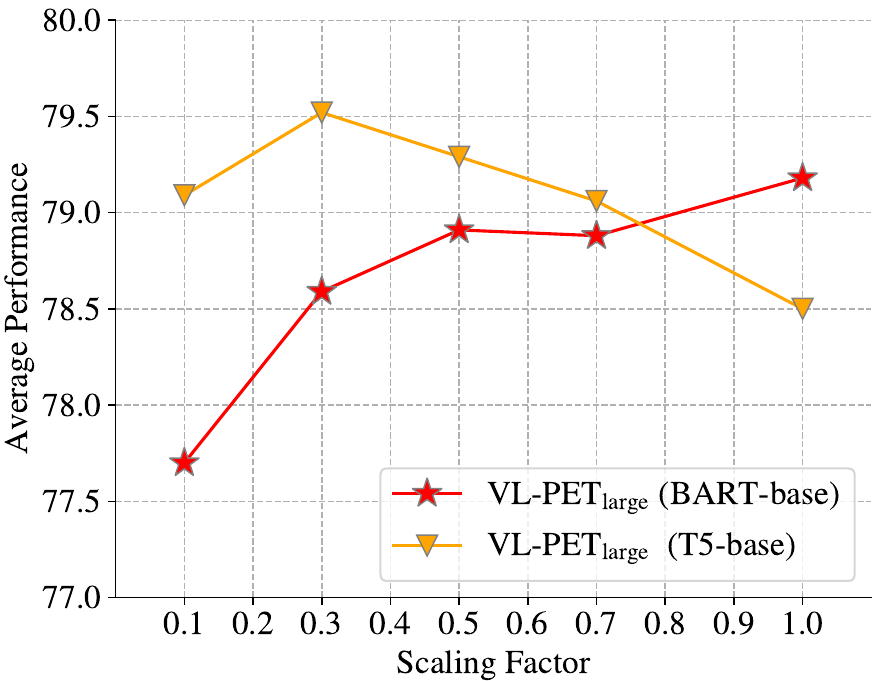}
\end{center}
   \caption{Effectiveness of scaling factor for the encoders. Experiments are conducted on image-text tasks based on different PLM backbones with one seed.}
\label{fig:BART_T5_scaling_factor}
\end{figure}

\section{Performance Improvement Effect of Granularity-controlled Mechanism}
Granularity control can dynamically assign importance weights to the intermediate hidden states. 
To fully investigate where the performance improvement is coming from, we replace the element-wise product with addition for $\mathbf{G}_\mathrm{large}$ on BART-base, denoted as VL-PET$_\mathrm{add}$. 
The results (VL-PET$_\mathrm{large}$ $>$ VL-PET$_\mathrm{add}$) in~\cref{tab:granularity_add} verify that the importance assignment is more effective than the added parameters in performance improvement.
Moreover, we observe that VL-PET$_\mathrm{add}$ introduces more trainable parameters but perform on par with VL-PET without granularity control, as their updating formulas are equivalent to conventional PET~(\cref{eq:unified}). 
These results further demonstrate the effectiveness of the proposed granularity control.

\section{Effectiveness of Scaling Factor}
As mentioned in~\cref{sec:granularity}, the scaling factor is specialized for different PLMs. 
Our experimental results in~\cref{fig:BART_T5_scaling_factor} show that the optimal scaling factors for BART-base encoders and T5-base encoders are 1.0 and 0.3, respectively.

\section{Unified Text Generation with Task Prompts}
Inspired by~\cite{Sung2021VLAdapter,Sung2022LST}, we prepend a task-specific prompt to the input sentence for each downstream task. The detailed task prompts are listed in~\cref{table:task_prompt}. We conduct experiments to test the validity of these task prompts over three seeds in~\cref{tab:task prompt}. For the BART-base backbone, all VL-PET modules with task prompts outperform those without task prompts. However, for T5-base, some VL-PET modules without task prompts exhibit superior performance compared to those with task prompts. 
For a fair comparison with state-of-the-art PET techniques, we still adopt task prompts to facilitate multi-task learning via unified text generation for our generative PLM backbones.

\begin{table*}[!t]
\begin{center}
    \resizebox{0.95\textwidth}{!}{
    \begin{tabular}{l c c c c c c}
    \toprule
    \makecell{Method} & \makecell{Trainable\\Params (\%)} &  \makecell{VQA\\Acc. (\%)} & \makecell{GQA\\Acc. (\%)} & \makecell{NLVR$^{2}$\\Acc. (\%)} & \makecell{COCO\\ Cap. (CIDEr)} &  \makecell{Avg.}\\  
    \midrule
    \midrule
    $\textbf{Backbone: BART-base}$ \\
    \midrule
\textbf{Without task prompts} \\
    VL-PET$_\mathrm{small}$   & 2.98 & 64.83$_{0.09}$ & 54.23$_{0.26}$ & 72.27$_{0.06}$ & 121.03$_{0.28}$ & 78.09$_{0.10}$ \\
    VL-PET$_\mathrm{middleX}$   & 2.98 & 65.14$_{0.16}$ & 54.55$_{0.09}$ & 72.77$_{0.27}$ & 120.91$_{0.33}$ & 78.34$_{0.07}$ \\
    VL-PET$_\mathrm{middleY}$  & 2.98 & 64.69$_{0.14}$ & 53.40$_{0.35}$ & 73.04$_{0.16}$ & 120.14$_{0.58}$ & 77.82$_{0.12}$ \\
    VL-PET$_\mathrm{large}$   & 4.16 & 65.78$_{0.08}$ & 54.45$_{0.32}$ & 72.90$_{0.36}$ & 121.46$_{0.36}$ & 78.65$_{0.22}$ \\

\textbf{With task prompts} \\
    VL-PET$_\mathrm{small}$   & 2.98 & 65.43$_{0.06}$ & 54.03$_{0.14}$ & 72.43$_{0.22}$ & 120.68$_{0.35}$ & 78.14$_{0.11}$ \\
    VL-PET$_\mathrm{middleX}$   & 2.98 & 65.54$_{0.09}$ & 54.53$_{0.15}$ & 72.66$_{0.17}$ & {120.72$_{0.51}$} & {78.37$_{0.14}$} \\
    VL-PET$_\mathrm{middleY}$   & 2.98 & 65.36$_{0.15}$ & 53.83$_{0.39}$ & {73.43$_{0.78}$} & 120.31$_{0.09}$ & 78.23$_{0.19}$ \\
    VL-PET$_\mathrm{large}$   & 4.16 & {66.17$_{0.27}$} & {55.11$_{0.17}$} & {73.43$_{0.35}$} & {122.03$_{0.46}$} & {79.18$_{0.14}$} \\
    \midrule
    \midrule
    $\textbf{Backbone: T5-base}$ \\
    \midrule
\textbf{Without task prompts} \\
    VL-PET$_\mathrm{small}$   & 4.51 & 65.67$_{0.31}$ & 56.09$_{0.29}$ & 73.54$_{0.63}$ & 120.51$_{1.11}$ & 78.96$_{0.57}$ \\
    VL-PET$_\mathrm{middleX}$   & 4.50 & 65.68$_{0.17}$ & 56.57$_{0.10}$ & 74.05$_{0.25}$ & 119.92$_{0.65}$ & 79.06$_{0.15}$ \\
    VL-PET$_\mathrm{middleY}$   & 4.50 & 65.75$_{0.34}$ & 56.35$_{0.41}$ & 73.93$_{0.60}$ & 119.84$_{1.08}$ & 78.97$_{0.28}$ \\
    VL-PET$_\mathrm{large}$   & 7.31 & 66.31$_{0.06}$ & 56.56$_{0.27}$ & 73.61$_{0.22}$ & 121.95$_{0.09}$ & 79.61$_{0.08}$ \\

\textbf{With task prompts} \\
VL-PET$_\mathrm{small}$   & 4.51 & 65.88$_{0.31}$ & 54.96$_{1.01}$ & 72.64$_{0.09}$ & 120.05$_{0.41}$ & 78.38$_{0.37}$ \\
VL-PET$_\mathrm{middleX}$ & 4.50  & 66.63$_{0.14}$ & 55.87$_{0.25}$ & {74.11$_{0.37}$} & {120.41$_{0.31}$} & {79.26$_{0.26}$} \\
VL-PET$_\mathrm{middleY}$ & 4.50  & 66.62$_{0.20}$ & 55.87$_{0.13}$ & 73.91$_{0.45}$ & 120.26$_{0.40}$ & 79.17$_{0.08}$ \\
VL-PET$_\mathrm{large}$   & 7.31 & {66.95$_{0.21}$} & {56.06$_{0.21}$} & 73.42$_{0.46}$ & {121.66$_{0.06}$} & {79.52$_{0.21}$} \\
    \bottomrule
    \end{tabular}
    }
    \end{center}
    \caption{Effectiveness of task prompts.}
    \label{tab:task prompt}
\end{table*}

\begin{table*}[!t]
\begin{center}
\resizebox{0.95\textwidth}{!}{
    \begin{tabular}{lcccccc}
    \toprule
    \makecell{Method} & \makecell{Params (\%)} &  \makecell{VQA (\%)} & \makecell{GQA (\%)} & \makecell{NLVR$^{2}$ (\%)} & \makecell{COCO (CIDEr)} &  \makecell{Avg.}\\  

    \midrule
VL-PET$_\mathrm{small}$ ($r=96$)
 & 2.98 & 65.43$_{0.06}$ & 54.03$_{0.14}$ & 72.43$_{0.22}$ & 120.68$_{0.35}$ & 78.14$_{0.11}$ \\
$r=144$                   & 3.58 & 65.87$_{0.12}$ & 54.05$_{0.21}$ & 72.76$_{0.09}$ & 121.22$_{0.40}$  & 78.48$_{0.07}$ \\
$r=192$                    & 4.16 & 66.14$_{0.16}$ & 54.73$_{0.18}$ & 72.63$_{0.23}$ & 121.46$_{1.38}$ & 78.74$_{0.34}$ \\
\midrule
VL-PET$_\mathrm{middleX}$ ($r=96$)  & 2.98 & 65.54$_{0.09}$ & 54.53$_{0.15}$ & 72.66$_{0.17}$ & 120.72$_{0.51}$ & 78.37$_{0.14}$ \\
$r=144$                    & 3.57 & 66.08$_{0.10}$ & 54.82$_{0.29}$ & 72.82$_{0.23}$ & 121.05$_{0.11}$ & 78.70$_{0.11}$ \\
$r=192$                   & 4.16 & 66.23$_{0.06}$ & 54.55$_{0.33}$ & 73.57$_{0.34}$ & 122.01$_{0.65}$ & 79.09$_{0.23}$ \\
\midrule
VL-PET$_\mathrm{middleY}$ ($r=96$) & 2.98 & 65.36$_{0.15}$ & 53.83$_{0.39}$ & 73.43$_{0.78}$ & 120.31$_{0.09}$ & 78.23$_{0.19}$ \\
$r=144$                   & 3.57 & 65.58$_{0.18}$ & 54.06$_{0.32}$ & 73.02$_{0.52}$ & 120.59$_{0.26}$ & 78.31$_{0.30}$ \\
$r=192$                   & 4.16 & 65.94$_{0.06}$ & 54.29$_{0.40}$ & 73.53$_{0.25}$ & 120.89$_{0.30}$ & 78.66$_{0.18}$ \\
\midrule
VL-PET$_\mathrm{large}$ ($r=96$) & 4.16 & 66.17$_{0.27}$ & 55.11$_{0.17}$ & 73.43$_{0.35}$ & 122.03$_{0.46}$ & 79.18$_{0.14}$ \\
$r=144$             & 5.31 & 66.47$_{0.10}$ & 55.07$_{0.25}$ & 73.21$_{0.15}$ & 122.05$_{0.58}$ & 79.20$_{0.13}$ \\
$r=192$             & 6.43 & 66.72$_{0.06}$ & 54.61$_{0.31}$ & 73.55$_{0.88}$ & 122.16$_{0.18}$ & 79.26$_{0.12}$ \\
    \bottomrule
    \end{tabular}
    }
    \end{center}
    \caption{Scalability of VL-PET Modules. ($r$: the projected hidden dimension of the encoder VL-PET modules.)}
    \label{tab:Scalable}
\end{table*}

\section{Scalability of VL-PET Modules}
We scale up the trainable parameters of the VL-PET modules by increasing the projected hidden dimensions $r$ of the Encoder VL-PET modules in BART-base. 
All scaled VL-PET modules outperform the unscaled ones in~\cref{tab:Scalable}, indicating the scalability of our VL-PET framework.

\begin{table*}[t]
\begin{center}
    \resizebox{0.95\textwidth}{!}{
    \begin{tabular}{l c c c c c c}
    \toprule
    \makecell{Method} & \makecell{Trainable\\Params (\%)} &  \makecell{VQA\\Acc. (\%)} & \makecell{GQA\\Acc. (\%)} & \makecell{NLVR$^{2}$\\Acc. (\%)} & \makecell{COCO\\ Cap. (CIDEr)} &  \makecell{Avg.}\\  
    \midrule
    \midrule
    $\textbf{Backbone: BART-base}$ \\
    \midrule
\textbf{Random Gaussian Initialization (Default)} \\
    VL-PET$_\mathrm{small}$   & 2.98 & 65.43$_{0.06}$ & 54.03$_{0.14}$ & 72.43$_{0.22}$ & 120.68$_{0.35}$ & 78.14$_{0.11}$ \\
    VL-PET$_\mathrm{middleX}$   & 2.98 & 65.54$_{0.09}$ & 54.53$_{0.15}$ & 72.66$_{0.17}$ & {120.72$_{0.51}$} & {78.37$_{0.14}$} \\
    VL-PET$_\mathrm{middleY}$   & 2.98 & 65.36$_{0.15}$ & 53.83$_{0.39}$ & {73.43$_{0.78}$} & 120.31$_{0.09}$ & 78.23$_{0.19}$ \\
    VL-PET$_\mathrm{large}$   & 4.16 & {66.17$_{0.27}$} & {55.11$_{0.17}$} & {73.43$_{0.35}$} & {122.03$_{0.46}$} & {79.18$_{0.14}$} \\

\textbf{Zero Initialization} \\
    VL-PET$_\mathrm{small}$   & 2.98 & 65.33$_{0.21}$ & 53.96$_{0.15}$ & 72.48$_{0.10}$ & 121.05$_{0.28}$ & 78.21$_{0.13}$ \\
    VL-PET$_\mathrm{middleX}$   & 2.98 & 65.49$_{0.06}$ & 54.28$_{0.20}$ & 73.01$_{0.47}$ & {120.36$_{0.26}$} & {78.29$_{0.15}$} \\
    VL-PET$_\mathrm{middleY}$  & 2.98 & 65.21$_{0.20}$ & 53.86$_{0.53}$ & 72.84$_{1.21}$ & 120.38$_{0.65}$ & 78.07$_{0.30}$ \\
    VL-PET$_\mathrm{large}$   & 4.16 & 66.18$_{0.13}$ & 54.44$_{0.58}$ & 72.99$_{0.69}$ & 121.91$_{0.52}$ & 78.88$_{0.35}$ \\
    \midrule
    \midrule
    $\textbf{Backbone: T5-base}$ \\
    \midrule
\textbf{Random Gaussian Initialization} \\
    VL-PET$_\mathrm{small}$   & 4.51 & 66.80$_{0.22}$ & 56.15$_{0.22}$ & 74.25$_{0.59}$ & 120.57$_{0.81}$ & 79.44$_{0.45}$ \\
    VL-PET$_\mathrm{middleX}$   & 4.50 & 66.33$_{0.47}$ & 55.89$_{0.05}$ & 74.23$_{0.21}$ & 119.79$_{0.59}$ & 79.06$_{0.31}$ \\
    VL-PET$_\mathrm{middleY}$   & 4.50 & 66.52$_{0.12}$ & 55.52$_{0.65}$ & 73.96$_{0.45}$ & 120.99$_{0.21}$ & 79.25$_{0.05}$ \\
    VL-PET$_\mathrm{large}$   & 7.31 & 66.78$_{0.10}$ & 55.62$_{0.18}$ & 73.21$_{0.15}$ & 121.12$_{0.36}$ & 79.19$_{0.08}$ \\

\textbf{Zero Initialization (Default)} \\
VL-PET$_\mathrm{small}$   & 4.51 & 65.88$_{0.31}$ & 54.96$_{1.01}$ & 72.64$_{0.09}$ & 120.05$_{0.41}$ & 78.38$_{0.37}$ \\
VL-PET$_\mathrm{middleX}$ & 4.50  & 66.63$_{0.14}$ & 55.87$_{0.25}$ & {74.11$_{0.37}$} & {120.41$_{0.31}$} & {79.26$_{0.26}$} \\
VL-PET$_\mathrm{middleY}$ & 4.50  & 66.62$_{0.20}$ & 55.87$_{0.13}$ & 73.91$_{0.45}$ & 120.26$_{0.40}$ & 79.17$_{0.08}$ \\
VL-PET$_\mathrm{large}$   & 7.31 & {66.95$_{0.21}$} & {56.06$_{0.21}$} & 73.42$_{0.46}$ & {121.66$_{0.06}$} & {79.52$_{0.21}$} \\
    \bottomrule
    \end{tabular}
    }
    \end{center}
    \caption{Experimental results of different weight initialization strategies.}
    \label{tab:weight}
\end{table*}

\section{Weight Initialization}
In~\cref{tab:weight}, we show the experimental results over three seeds with two popular weight initialization strategies, i.e., random Gaussian initialization and zero initialization, for both BART-base and T5-base backbones. 
Random Gaussian initialization is widely used in PET techniques to initialize the weight of PET modules from a Gaussian distribution, 
while some PET techniques~\cite{hu2021lora} utilize zero initialization to set the up projection layers of the PET modules as zero.
Based on~\cref{tab:weight}, it can be inferred that random Gaussian initialization is more appropriate for BART-base and zero initialization for T5-base. 
Therefore, we employ random Gaussian initialization for BART-base and zero initialization for T5-base by default in this work.

\section{More Designs for Multi-head Modular Modification}
We propose a multi-head modular modification $\Delta{\mathbf{H}^\prime} \in \mathbb{R}^{N \times d}$ of length $N$ and dimension $d$ in the main paper, which is formulated as follows:
\begin{equation}
    \Delta{\mathbf{H}^\prime} 
    =  \phi(\text{Concat}(
    \mathbf{X}^\prime{\mathbf{W}_\text{down}^{(1)}}, \cdots, \mathbf{X}^\prime\mathbf{W}_\text{down}^{(N_h)}))
    \mathbf{W}_\text{up},
\end{equation}
where $N_h$ is the number of heads, 
$\mathbf{X}^\prime \in \mathbb{R}^{N \times d}$ is the input, $\mathbf{W}_\text{down}^{(i)}\in \mathbb{R}^{d \times \frac{r}{N_h}}$ is a down projection layer for the $i$-th head,
$\phi$ is the GELU function~\cite{hendrycks2016gaussian}, $\mathbf{W}_\text{up} \in \mathbb{R}^{r \times d}$ is a up projection layer and $r$ is the projected hidden dimension. 

Since the multi-head idea is applied to the down projection layer in the main paper, we can call this type of multi-head modular modification as down multi-head modular modification.
As the multi-head idea can be applied to any linear projection within a modular modification, it yields a series of variant designs. 
In this section, we present additional designs of multi-head modular modification, such as up multi-head modular modification, down-up multi-head modular modification and down-up-pair multi-head modular modification. The primary difference among these four designs lies in where the multi-head idea is employed.

For up multi-head modular modification, we apply the multi-head idea to the up projection layer and describe it as follows:
\begin{equation}
\begin{split}
    \Delta{\mathbf{H}^\prime} 
    =  \text{Concat}(&\phi((\mathbf{X}^\prime{\mathbf{W}_\text{down}}))
    \mathbf{W}_\text{up}^{(1)}, \cdots,\\
    &\phi((\mathbf{X}^\prime{\mathbf{W}_\text{down}}))
    \mathbf{W}_\text{up}^{(N_h)}),
\end{split}
\end{equation}
where $\mathbf{W}_\text{down} \in \mathbb{R}^{d \times r}$ is a down projection layer and $\mathbf{W}_\text{up}^{(i)}\in \mathbb{R}^{r \times \frac{d}{N_h}}$ is a down projection layer for head$_i$.

\begin{table*}[t]
\begin{center}
    \resizebox{0.95\textwidth}{!}{
    \begin{tabular}{l c c c c c c}
    \toprule
    \makecell{Method} & \makecell{Trainable\\Params (\%)} &  \makecell{VQA\\Acc. (\%)} & \makecell{GQA\\Acc. (\%)} & \makecell{NLVR$^{2}$\\Acc. (\%)} & \makecell{COCO\\ Cap. (CIDEr)} &  \makecell{Avg.}\\  
    \midrule
    \midrule
    $\textbf{Backbone: BART-base}$ \\
    $\textbf{Granularity-controlled Mechanism:} \mathbf{G}_\mathrm{large}$ \\
    \midrule
\textbf{Down Multi-head Modular Modifications} \\
(used in the main paper)\\
    $N_h = 2$   & 4.16 & 66.20$_{0.12}$ & 54.74$_{0.43}$ & 72.88$_{0.43}$ & 121.53$_{0.64}$ & 78.84$_{0.26}$ \\
    $N_h = 4$   & 4.16 & 66.17$_{0.27}$ & 55.11$_{0.17}$ & 73.43$_{0.35}$ & 122.03$_{0.46}$ & 79.18$_{0.14}$ \\
\textbf{Up Multi-head Modular Modifications} \\
    $N_h = 2$   & 4.16 & 66.16$_{0.12}$ & 54.88$_{0.21}$ & 73.19$_{0.13}$ & 121.69$_{0.62}$ & 78.98$_{0.14}$ \\
    $N_h = 4$   & 4.16 & 66.15$_{0.13}$ & 55.00$_{0.36}$ & 73.19$_{0.27}$ & 121.44$_{0.73}$ & 78.95$_{0.23}$ \\
\textbf{Down-Up Multi-head Modular Modifications} \\
    $N_h = 2$   & 4.16 & 66.24$_{0.10}$ & 54.73$_{0.24}$ & 72.81$_{0.21}$ & 121.56$_{0.15}$ & 78.84$_{0.07}$ \\
    $N_h = 4$   & 4.16 & 66.18$_{0.16}$ & 54.60$_{0.27}$ & 73.17$_{0.42}$ & 122.05$_{0.20}$ & 79.00$_{0.10}$ \\
\textbf{Down-Up-Pair Multi-head Modular Modifications} \\
    $N_h = 2$   & 3.86 & 65.86$_{0.04}$ & 54.51$_{0.36}$ & 72.98$_{0.19}$ & 121.65$_{0.21}$ & 78.75$_{0.07}$ \\
    $N_h = 4$   & 3.72 & 65.70$_{0.09}$ & 54.24$_{0.12}$ & 72.43$_{0.35}$ & 122.76$_{0.33}$ & 78.28$_{0.20}$ \\
    \bottomrule
    \end{tabular}
    }
    \end{center}
    \caption{Experimental results of multi-head modular modifications with various numbers of heads.}
    \label{tab:modifications}
\end{table*}

For down-up multi-head modular modification, we apply the multi-head idea to the down projection layer and up projection layer, respectively. The formulation is described as follows:
\begin{equation}
\begin{split}
    &\Delta{\mathbf{H}^\prime}_\text{down} 
    =  \phi(\text{Concat}(
    \mathbf{X}^\prime{\mathbf{W}_\text{down}^{(1)}}, \cdots, \mathbf{X}^\prime\mathbf{W}_\text{down}^{(N_h)})), \\
    &\Delta{\mathbf{H}^\prime} 
    =  \text{Concat}(
    \Delta{\mathbf{H}^\prime}_\text{down}
    \mathbf{W}_\text{up}^{(1)}, \cdots,
    \Delta{\mathbf{H}^\prime}_\text{down}
    \mathbf{W}_\text{up}^{(N_h)}),
\end{split}
\end{equation}
where $\Delta{\mathbf{H}^\prime}_\text{down} \in \mathbb{R}^{N \times r}$ is an intermediate output.

For down-up-pair multi-head modular modification, we formulate it as follows:
\begin{equation}
\begin{split}
    &\Delta{\mathbf{H}^\prime}^{(i)}_\text{pair}
    =  \phi(\mathbf{X}^\prime{\mathbf{W}_\text{down-pair}^{(i)}})\mathbf{W}_\text{up-pair}^{(i)}, \\
    &\Delta{\mathbf{H}^\prime} 
    =  \text{Concat}(
    \Delta{\mathbf{H}^\prime}^{(1)}_\text{pair}, \cdots,
    \Delta{\mathbf{H}^\prime}^{(N_h)}_\text{pair}),
\end{split}
\end{equation}
where 
$\mathbf{W}_\text{down-pair}^{(i)}\in \mathbb{R}^{d \times \frac{r}{N_h}}$ is a down projection layer, $\mathbf{W}_\text{up-pair}^{(i)}\in \mathbb{R}^{\frac{r}{N_h} \times \frac{d}{N_h}}$ is a up projection layer 
and $\Delta{\mathbf{H}^\prime}^{(i)}_\text{pair} \in \mathbb{R}^{N \times \frac{d}{N_h}}$ is an intermediate output for head$_i$.

Similar to the usage of multi-head modular modifications in the main paper, we perform experiments on image-text tasks with BART-base backbone for these four multi-head modular modifications with our proposed granularity-controlled mechanism at a large level. 

As shown in~\cref{tab:modifications}, the down multi-head modular modification with $N_h=4$ achieves the best performance in the experiment, which is adopted in the main paper.
We observe that even the worst down-up-pair multi-head modular modification still outperforms the state-of-the-art PET techniques (e.g., VL-Adapter). 
Moreover, down-up-pair multi-head modular modification significantly reduces the number of trainable parameters as the number of heads increases, indicating a promising direction for further parameter reduction with a minimal impact on performance.

\section{Limitations}
In this section, we discuss some limitations of our work.
\begin{itemize}
  \item We perform extensive experiments and analysis on image-text tasks and video-text tasks. 
However, the video-text experiments are conducted with only one seed due to the submission limit of the VALUE benchmark, which may affect the reliability of the video-text experimental results. 
  \item Our VL-PET framework focuses on challenging VL downstream tasks, including some discriminative and generative tasks (e.g., question-answering tasks and captioning tasks).
But there are many other downstream tasks in the real life, and we do not include all types of tasks in the multi-task learning experiments. 
Therefore, our designs and experimental results are not always guaranteed to be generalized to other VL tasks (e.g., image-text retrieval and video-text retrieval) and other domains (e.g., NLP and CV).
  \item We validate the enhanced effect of employing our VL-PET designs (e.g., granularity-controlled mechanism and lightweight PET module designs) on existing PET techniques (i.e., Compacter and VL-Adapter). 
But our VL-PET designs may not be applicable to enhance all PET techniques.
    \item PET techniques are proposed to reduce the number of trainable parameters and save model storage space. Our work utilizes multi-task learning to achieve further reduction.
However, similar to most PET techniques, our work integrates newly-designed modules into large PLM backbones, which still evoke a lot of memory and time consumption during training and inference.

\end{itemize}

\end{document}